\documentclass[a4paper, 10pt, conference]{ieeeconf}      % Use this line for a4
\usepackage{FG2026}
\usepackage{multirow, graphicx}
\usepackage[dvipsnames]{xcolor}
\usepackage{float}
\usepackage{hyperref}
\usepackage{breakurl} 
\usepackage{xurl}

\FGfinalcopy % *** Uncomment this line for the final submission

\IEEEoverridecommandlockouts                              % This command is only
\def\FGPaperID{****} % *** Enter the FG2026 Paper ID here

\title{\LARGE \bf
On the Impact of Face Segmentation-Based Background Removal on Recognition and Morphing Attack Detection
}

\author{\parbox{16cm}{\centering
    {\large Eduarda Caldeira$^{1,2}$, Guray Ozgur$^{1,2}$, Fadi Boutros$^{1}$ and Naser Damer$^{1,2}$}\\
    {\normalsize
    $^1$Fraunhofer Institute for Computer Graphics Research IGD, Germany\\
    $^2$Department of Computer Science, TU Darmstadt, Germany}}
    \thanks{This research work has been funded by the German Federal Ministry of Education and Research and the Hessian Ministry of Higher Education, Research, Science and the Arts within their joint support of the National Research Center for Applied Cybersecurity ATHENE.}% <-this % stops a space
}
\usepackage{fancyhdr}
\begin{document}

\ifFGfinal
\thispagestyle{empty}
\pagestyle{empty}
\else

\pagestyle{plain}
\fi
\maketitle
\thispagestyle{fancy}

\begin{abstract}
This study investigates the impact of face image background correction through segmentation on face recognition and morphing attack detection performance in realistic, unconstrained image capture scenarios. The motivation is driven by operational biometric systems such as the European Entry/Exit System (EES), which require facial enrolment at airports and other border crossing points where controlled backgrounds usually required for such captures cannot always be guaranteed, as well as by accessibility needs that may necessitate image capture outside traditional office environments. By analyzing how such preprocessing steps influence both recognition accuracy and security mechanisms, this work addresses a critical gap between usability-driven image normalization and the reliability requirements of large-scale biometric identification systems. Our study evaluates a comprehensive range of segmentation techniques, three families of morphing attack detection methods, and four distinct face recognition models, using databases that include both controlled and in-the-wild image captures. The results reveal consistent patterns linking segmentation to both recognition performance and face image quality. Additionally, segmentation is shown to systematically influence morphing attack detection performance. These findings highlight the need for careful consideration when deploying such preprocessing techniques in operational biometric systems. \url{https://github.com/EduardaCaldeira/FSB-BR}
\end{abstract}    
\section{Introduction}
\label{sec:intro}

\begin{figure}
    \centering
    \includegraphics[width=\linewidth]{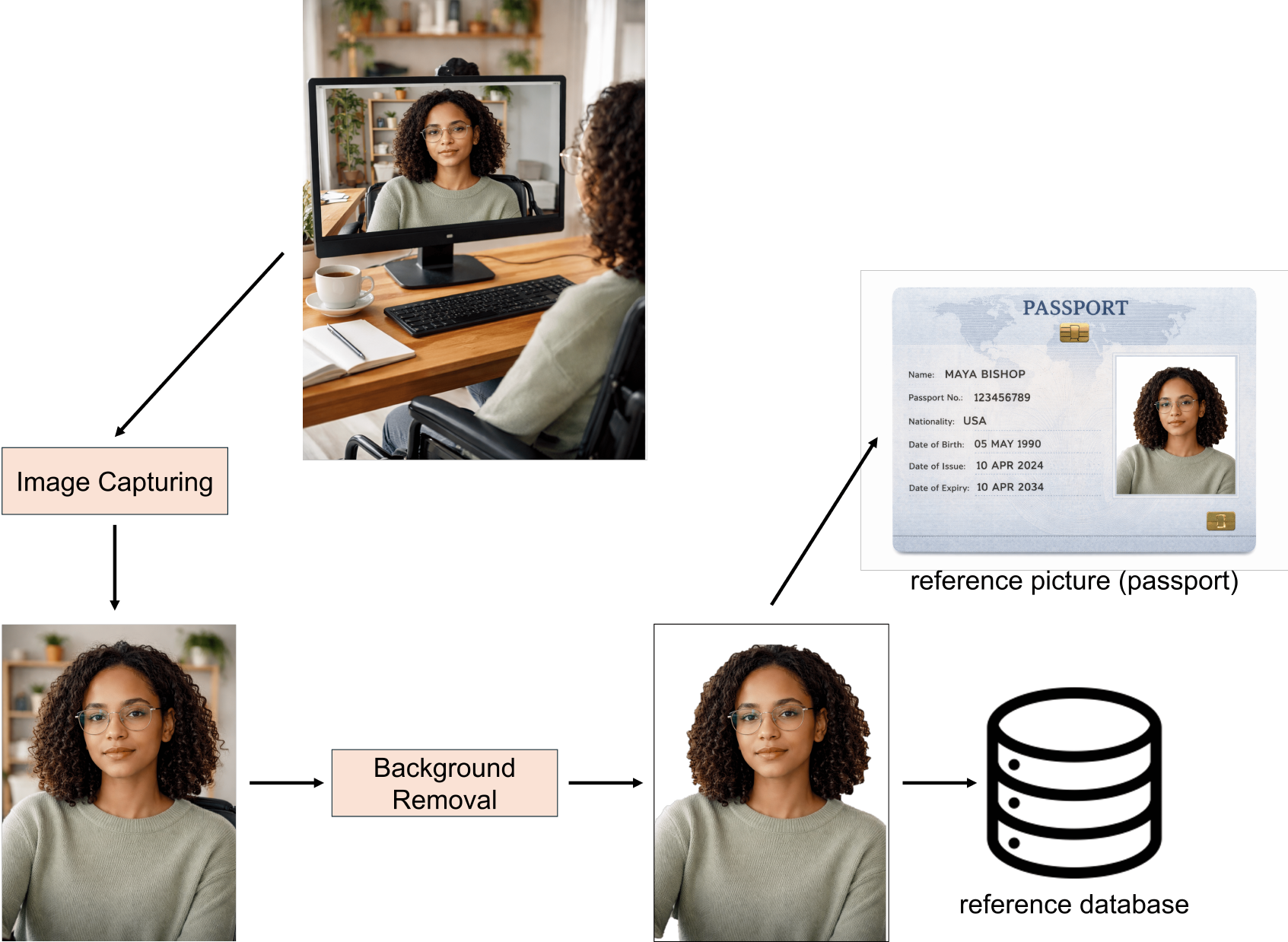}
    \vspace{-5mm}
    \caption{Visual example of this work's use case. The image can be acquired in a non-standard environment, facilitating acquisition for individuals with low mobility. After background removal, it is stored in the EES system database and can be used as a reference in future border control processes.}
    \vspace{-6mm}
    \label{fig:pipeline}
\end{figure}

The increasing deployment of biometric systems in large-scale identity management and border control has intensified the need for reliable face image acquisition under unconstrained conditions. A prominent example is the European Entry/Exit System (EES) \cite{orav2016smart}, which requires the enrollment of travelers’ biometric data, including facial images, at airports and other border crossing points upon arrival. In such high-throughput and heterogeneous environments, ensuring controlled capture conditions, such as uniform backgrounds, stable illumination, and optimal camera positioning, is often impractical. These challenges are further amplified when enrollment is extended beyond traditional offices to self-service kiosks or assisted capture scenarios. In addition, inclusive design requirements may necessitate image acquisition in non-standard environments, such as private or semi-private spaces, for individuals with limited mobility (Figure \ref{fig:pipeline}).

\textcolor{black}{To guarantee reliable recognition performance, large-scale biometric systems require high-quality enrollment images. This need is formalized through standards such as ISO/IEC 19794-5 \cite{iso19794-5_2011}, which define strict requirements on facial image properties. In particular, although modern face recognition systems typically rely on landmark-based cropping to isolate the face region before recognition, ISO standards still impose image requirements that include background constraints, such as the presence of a plain, light-colored background}. However, such constraints are difficult to satisfy in the aforementioned real-world scenarios. As a result, face segmentation and background correction \cite{adobe_express_bg_removal, removebg, backgrounderaser_app} have emerged as practical solutions to transform unconstrained captures into standardized, ISO-compliant images, at least in terms of unifying the background. By removing complex or cluttered backgrounds, these techniques aim to improve interoperability and maintain recognition performance without imposing restrictive acquisition setups.

Despite their practical appeal, these preprocessing steps fundamentally modify the original image content, raising an important question: to what extent do they affect the reliability and security of biometric systems? Face recognition (FR) algorithms are known to be sensitive to changes in facial texture, contours, and contrast, particularly in regions encoding identity-discriminative information \cite{DBLP:conf/icpr/DamerWHCB0K18, lee2020byeglassesgan, zeng2021image, rathgeb2019impact}. Simultaneously, security mechanisms such as morphing attack detection (MAD), which aim to detect fraudulent enrollment attempts \cite{DBLP:conf/icb/FerraraFM14,iso20059_2025}, rely on identifying subtle artifacts that may also be altered or suppressed by image preprocessing \cite{MADVgg, DBLP:conf/visapp/MakrushinND17, neddo2024printscanning, rathgeb2021compression}. In operational contexts such as airport enrollment, even minor degradations in recognition accuracy or increases in false alarms can lead to significant practical consequences, including delays and unnecessary secondary inspections.

Despite the possible and well-motivated adoption of segmentation-based background removal, its downstream impact on both recognition performance and attack detection remains insufficiently understood. Existing studies have primarily focused on isolated effects of image manipulation, without jointly analyzing their influence across the full biometric processing pipeline \cite{DBLP:conf/icpr/DamerWHCB0K18, fang2022unsupervised, neddo2024printscanning, rathgeb2019impact, zeng2021image}. A systematic evaluation that considers segmentation quality, face recognition, and morphing attack detection in an integrated manner is therefore essential. By analyzing these components jointly, this work addresses a critical gap between algorithmic development and real-world deployment, providing insights that can guide the design of robust preprocessing strategies and support the reliable operation of biometric systems under realistic conditions.

In this work, we systematically investigate the impact of face segmentation-based background removal on the full biometric processing pipeline, addressing the lack of joint analysis identified in prior work, which has largely focused on isolated effects of image manipulation on either FR \cite{DBLP:conf/icpr/DamerWHCB0K18, zeng2021image} or MAD \cite{fang2022unsupervised, neddo2024printscanning}, but not both simultaneously, nor specifically in the context of background removal for ISO-compliant \cite{iso19794-5_2011} facial image correction. We evaluate seven representative segmentation techniques spanning classical, real-time, and foundation-model-based approaches, and analyze their effects across multiple operational scenarios using, controlled 
FERET \cite{phillips1998feret}, \textcolor{black}{semi-controlled FRGCv2 \cite{phillips2005overview}} and unconstrained IARPA Janus Benchmark–C (IJB-C) \cite{DBLP:conf/icb/MazeADKMO0NACG18}.
%(FERET) and unconstrained (IJB-C) datasets. 
Our experiments measure (i) face recognition performance through verification accuracy, (ii) shifts in similarity score distributions to capture subtle degradations, (iii) face image quality via face image quality assessment (FIQA) metrics, and (iv) MAD behavior in terms of false triggering rates (Bona fide Presentation Classification Error Rate (BPCER) at fixed thresholds). The results reveal that while segmentation has a negligible impact under controlled conditions, it introduces measurable degradation in recognition performance and image quality in unconstrained settings, with strong dependence on the segmentation method. Furthermore, segmentation systematically alters MAD behavior, reducing false triggers for some methods while significantly degrading others, particularly foundation-model-based approaches. Overall, our findings highlight a critical trade-off between usability-driven background normalization and the reliability of biometric systems, emphasizing that segmentation cannot be treated as a neutral preprocessing step and must be carefully evaluated in deployment-specific contexts.
\section{Related Work}
\label{sec:sota}
\subsection{Effects of Image Manipulation on FR and MAD}
\label{sec:manipulation} 

Image manipulation encompasses a broad range of processing operations that can affect both FR performance and MAD reliability. In the FR field, researchers have studied how common image processing operations degrade or alter recognition accuracy. Zeng~et~al. \cite{zeng2021image} examined the effects of denoising and image enhancement methods on deep FR, finding that image quality significantly impacts recognition precision and that Gaussian filtering outperforms self-snake model-based enhancement for this task. \cite{DBLP:conf/icpr/DamerWHCB0K18} investigated the robustness of deep learning-based FR to perspective distortion, showing that geometric deformations measurably degrade verification performance. The impact of facial beautification and retouching on FR systems has been surveyed by Rathgeb~et~al. \cite{rathgeb2019impact}, who provide a comprehensive overview of detection schemes and recognition vulnerabilities arising from cosmetic image edits. Regarding occlusion-induced degradation, Lee and Lai \cite{lee2020byeglassesgan} proposed ByeGlassesGAN, a GAN-based framework for identity-preserving eyeglass removal that, when applied as a pre-processing step, substantially improves FR accuracy for subjects wearing glasses. Schlett~et~al. \cite{schlett2023compression} further investigated the effect of lossy compression algorithms on face image quality and recognition, demonstrating that compression-induced degradation measurably affects verification performance. 

Within MAD literature, the influence of various image %post-
processing operations on the robustness of MAD algorithms has received growing attention. Fang~et~al. \cite{fang2022unsupervised} state that such variations can be related to image compression \cite{DBLP:conf/visapp/MakrushinND17}, the source of bona fide images \cite{DBLP:conf/iwbf/ScherhagRB18}, and image re-digitization \cite{MADVgg, DBLP:conf/btas/RaghavendraRB16,DBLP:conf/cvpr/RaghavendraRVB17a}. 
%Qin~et~al. \cite{DBLP:journals/tbbis/QinPVRLB21} proposed a partial face manipulation-based morphing attack targeting only the most identity-discriminative facial components, yielding morphs that are harder to detect and more visually similar to bona fide images. 
Rathgeb~et~al. \cite{rathgeb2021compression} studied the effects of JPEG and JPEG~2000 image compression on facial retouching detection, showing that even moderate compression levels can hamper manipulation detection algorithms. The effect of print-scanning on morphed images has also been investigated \cite{neddo2024printscanning}, revealing that subjecting morphed images to the print-scan process increases match rates against contributing subjects and degrades the performance of MAD algorithms across heterogeneous evaluation scenarios.

\subsection{Portrait Segmentation for Face Background Removal}
\label{sec:sota_seg} 

Background removal for face images is commonly implemented through semantic or prompt-based segmentation models that offer different trade-offs between boundary precision, robustness, and computational efficiency. Early fully convolutional formulations established dense pixel-wise prediction as a practical paradigm for segmentation \cite{long2015fully}, and lightweight FCN variants coupled with efficient backbones such as MobileNetV2 are widely used when fast inference is required. Multi-scale feature aggregation was further strengthened by Feature Pyramid Networks (FPN) \cite{lin2017feature}, which combine top-down and lateral connections to preserve semantically strong representations across resolutions. Real-time methods such as BiSeNet \cite{yu2018bisenet} and Fast-SCNN \cite{DBLP:conf/bmvc/PoudelLC19} explicitly separate high-resolution spatial detail from low-resolution contextual encoding to maintain accuracy under strict latency constraints. In parallel, context modeling based on attention has shown strong segmentation performance: DANet \cite{fu2019dual} introduces complementary position and channel attention modules to capture long-range dependencies in both spatial and feature dimensions.

More recent architectures improve generalization and efficiency by replacing heavy decoders or introducing foundation-model priors. SegFormer \cite{xie2021segformer} combines a hierarchical transformer encoder with a lightweight MLP decoder, removing positional encoding and achieving a strong accuracy-efficiency balance across model scales. Segment Anything (SAM) \cite{kirillov2023segment} extends segmentation to a promptable setting, where points, boxes, or masks can guide zero-shot object extraction across domains. This promptable behavior is particularly relevant for face background removal in unconstrained captures, where landmark- or box-guided prompting can better preserve facial regions while suppressing cluttered backgrounds. Outside academic pipelines, background removal is also frequently performed using consumer tools, including gallery editing features in mobile applications \cite{backgrounderaser_app} and online design platforms \cite{adobe_express_bg_removal} or services \cite{removebg}, which find their way into identity document applications, in some cases also being recommended by the officials unofficially. Despite the maturity of these methods and their wide use, their downstream impact on FR and MAD remains underexplored, motivating the comparative analysis conducted in this work.
\section{Methodology and Experimental Setup}
\label{sec:methodology}
%This section presents the methodology and experimental setup followed in this work. We first introduce the studied datasets and the segmentation methods used for background removal. Then, we provide a detailed explanation of the experimental protocol followed to assess the impact of background removal in FR, face image quality and MAD systems' inadequate triggering. 

\subsection{Datasets}

\textcolor{black}{We conduct the proposed study in three datasets: FERET \cite{phillips1998feret}, FRGCv2 \cite{phillips2005overview} and IARPA Janus Benchmark–C (IJB-C) \cite{DBLP:conf/icb/MazeADKMO0NACG18}. We specifically select them to provide results on data acquired under different conditions: a controlled setting with images of near passport-style characteristics (FERET's frontal images), a semi-controlled setting with frontal pictures acquired in both controlled and uncontrolled environments (FRGCv2), and an uncontrolled setting containing pictures with multiple headposes and backgrounds (IJB-C).}

\textbf{FERET:} The FERET dataset \cite{phillips1998feret} contains 11,338 facial images of size $256 \times 384$ from 994 subjects. The images were acquired over several photography sessions and encompass a wide range of facial positions (frontal view, right and left profile, quarter profile, and half profile). In this work, we filter FERET and consider only its frontal view samples, as this is the only head pose that aligns with the real ESS use case.  The frontal images were selected using the provided metadata by the FERET \cite{phillips1998feret} dataset, where each sample was labeled based on the pose.
%This filtering process is trivial, as each filename encodes the head pose information of the respective image. 
The 2,722 resultant images are divided into two categories: Frontal Gallery (FA) and Frontal Probe (FB), originally used as gallery and probe images, respectively \cite{phillips1998feret}. %Furthermore, images belonging to these two subsets present different facial expressions \cite{phillips1998feret}. 
Due to the absence of a standard face verification protocol for FERET, we generate a list of all genuine (pairs of images belonging to the same person) and impostor (pairs of images belonging to different people) combinations between FA and FB images, resulting in a total of 2506 genuine and 1,849,806 impostor pairs.

\textbf{FRGCv2:} \textcolor{black}{The FRGCv2 dataset \cite{phillips2005overview} was collected at the University of Notre Dame during two academic years, containing images captured in both controlled and uncontrolled conditions. Following the experimental protocols proposed in the original dataset release \cite{phillips2005overview}, we consider images taken in the fall and spring semesters of 2003 and the spring semester of 2004 for evaluation, resulting in a total of 35,276 images of size $1704 \times 2272$ and $2272 \times 1704$. However, since the original lists of pairs used in each experiment are no longer available, and given the need for a significant amount of comparisons to ensure that FR performance fluctuations are statistically significant, we design our own pairing protocol. After removing the 40 images for which no landmarks or boxes were detected during the alignment process \cite{DBLP:conf/cvpr/DengGVKZ20}, we generate a list of all genuine and impostor combinations, resulting in 1,782,172 genuine and over 600M impostor pairs. To reduce the computational overhead while maintaining a significant amount of comparisons, we randomly select 1\% of the impostor comparisons, resulting in 6,225,166 pairs.}

%image size (apparently there are two in the paper, but not in the dataset I have)

\textbf{IJB–C:} The IJB-C \cite{DBLP:conf/icb/MazeADKMO0NACG18} is a large-scale dataset consisting of 31,334 images of different sizes belonging to 3,531 distinct subjects. Following the original dataset release \cite{DBLP:conf/icb/MazeADKMO0NACG18}, we considered the official 1:1 mixed verification pairing protocol in our experiments, which contains 19,557 genuine and 15,638,932 impostor comparisons. 

\subsection{Background Removal Methods}
The background removal process was performed by substituting the background information of each image with white pixels. \textcolor{black}{To provide a diverse and representative evaluation of different architectural paradigms, we select segmentation methods spanning CNN-based and multi-scale feature fusion models \cite{lin2017feature, long2015fully}, attention-based architectures \cite{fu2019dual}, efficient real-time networks \cite{DBLP:conf/bmvc/PoudelLC19, yu2018bisenet}, transformer-based approaches \cite{xie2021segformer}, and promptable foundation models \cite{kirillov2023segment}.} The faces were segmented using seven different methods: FPN + ResNet50 \cite{lin2017feature}, SegFormer-B0 \cite{xie2021segformer}, BiSeNetv2 \cite{yu2018bisenet}, DANet \cite{fu2019dual}, Fast SCNN \cite{DBLP:conf/bmvc/PoudelLC19}, FCN + MobileNetv2 \cite{long2015fully}, and SAM \cite{kirillov2023segment}. \textcolor{black}{For the first six strategies, we used a version of each model pre-trained on the EasyPortrait \cite{kvanchiani2023easyportrait} dataset for the portrait segmentation task. Before segmentation, each image was resized to comply with the size each network was trained on ($224\times224$ for FPN + ResNet50 and SegFormer-B0, and $384\times384$ for the remaining systems)}. Each architecture outputs a mask separating the two test classes (background and person), which can be used for background removal. SAM \cite{kirillov2023segment} works differently, as it can receive input prompts that orient the model towards identifying the desired object, such as points of the image belonging (or not) to the target objects, or a bounding box that includes the desired object. In this work, we provide these two types of input prompts to aid SAM in the segmentation. \textcolor{black}{Since the samples contain one main object (the face), the perimeter of the image can be used to highlight that the whole face (instead of individual facial components) should be segmented. This technique is efficient for FERET, IJB-C and the portrait images of FRGCv2, since the face occupies most of the frame. However, FRGCv2 also contains landscape images, in which the background is predominant. For these samples, we extend the bounding boxes extracted by RetinaFace \cite{DBLP:conf/cvpr/DengGVKZ20} by 150\% to each side, 15\% vertically upward, and until the frame border vertically down. The empirical validation of these choices is provided in the supplementary material.
}We also consider the five facial landmarks extracted by RetinaFace \cite{DBLP:conf/cvpr/DengGVKZ20} as an extra orienting prompt. These points are defined as belonging to the sample, making it explicit that the whole face needs to be segmented, and highlighting important regions that may be occluded (such as the eyes, when glasses or sunglasses are present). \textcolor{black}{When segmenting with SAM, the images are fed in their original size, as the model automatically converts them and the accompanying prompts to the expected input size of $1024\times1024$, by rescaling the image and padding the shorter side \cite{kirillov2023segment}. All these segmentation networks are evaluated on an independent database in the supplementary material.}

\subsection{Face Recognition}
To assess the impact of the facial images' background removal on FR performance, the original and segmented datasets were evaluated by pre-trained FR models. To provide a thorough assessment, we test four different FR models in this step. We select two CNN-based (ElasticFace \cite{DBLP:conf/cvpr/BoutrosDKK22} and ArcFace \cite{DBLP:conf/cvpr/DengGXZ19}) and two ViT-based (SwinFace \cite{DBLP:journals/tcsv/QinWDWCHD24} and TransFace \cite{DBLP:conf/iccv/DanLXD0XS23}) models since the distinct ways CNNs and ViT networks process input samples might impact their response to image manipulation differently. ElasticFace uses a ResNet-100 \cite{he2016deep} architecture trained on MS1MV2 \cite{DBLP:conf/cvpr/DengGXZ19} with the ElasticFace-Cos \cite{DBLP:conf/cvpr/BoutrosDKK22} loss function. ArcFace follows a ResNet-100 architecture trained on MS1MV3 \cite{deng2019lightweight} with the ArcFace \cite{DBLP:conf/cvpr/DengGXZ19} loss. SwinFace \cite{DBLP:journals/tcsv/QinWDWCHD24} used the MS1MV2 \cite{DBLP:conf/cvpr/DengGXZ19} dataset to train a multi-task transformer capable of performing face recognition, expression recognition, age estimation, and attribute estimation. In the current work, only the identity branch was used during evaluation. TransFace \cite{DBLP:conf/iccv/DanLXD0XS23} is a vision transformer architecture trained for the FR task with data augmentation and hard sample mining strategies specifically designed to preserve face structural information and leverage local tokens' information, enhancing the transformer's FR performance. In this study, TransFace-L trained on Glint360k \cite{an2022killing} was selected for FR evaluation, given its superior FR performance compared with the remaining available pre-trained TransFace models \cite{DBLP:conf/iccv/DanLXD0XS23}. Before being fed to these models, the evaluated samples were aligned with RetinaFace \cite{DBLP:conf/cvpr/DengGVKZ20} and resized to $112 \times 112$ to comply with their input requirements. \textcolor{black}{Due to their large dimensions, FRGCv2 images were rescaled so that their largest size had the maximum size of $1908$ pixels \cite{DBLP:conf/cvpr/DengGVKZ20}, to allow seamless landmark and box detection.}

The FR verification performances are evaluated and reported as the verification accuracy achieved by each FR system on each version of FERET, and FRGCv2, following the previously described pairing protocols. However, as shown in the results section, the FR performances achieved by these datasets are not discriminative (very high FR verification accuracies on frontal images) and thus do not allow for drawing clear conclusions regarding the impact of background removal techniques on their own. \textcolor{black}{Hence, the verification accuracy is complemented with three extra metrics, namely the average cosine similarity of genuine ($gen_{avg}$) and impostor ($imp_{avg}$) pairs, and the difference between them ($\Delta$).} These metrics can highlight slight distribution shifts resultant from background removal that are not sufficient to significantly change the overlapping region between genuine and impostor distributions, and are thus not detectable through the sole analysis of FR verification accuracy. For the large-scale benchmark IJB-C, we reported the verification performance as True Acceptance Rates (TAR) at False Acceptance Rates (FAR) of 1e-4 and 1e-5 \cite{DBLP:conf/icb/MazeADKMO0NACG18}, \textcolor{black}{along with $gen_{avg}$, $imp_{avg}$, and $\Delta$}.

\subsection{Face Image Quality Assessment (FIQA)}
We further analyze whether background removal impacts the facial image quality from the FR perspective. We employ CR-FIQA(L) \cite{boutros2023cr}, a state-of-the-art FIQA solution that achieves the highest performance in several aspects of the NIST Face Analysis Technology Evaluation (FATE) Quality \cite{NISTQuaity}. We extract the average and standard deviation values of the output FIQA metric for each dataset before and after segmentation with each of the considered methods.

\subsection{Morphing Attack Detection}
To verify whether removing the background of facial images acquired by the EES would affect how often MAD systems are triggered with false positives, the original and segmented datasets were subjected to three different morphing attack detection protocols. SPL \cite{fang2022unsupervised} is an unsupervised face morphing attack detection method based on anomaly detection. It was trained on CASIA-WebFace \cite{DBLP:journals/corr/YiLLL14a} and SMDD \cite{damer2022privacy} with a self-paced learning strategy that assigns smaller weights to suspicious samples, allowing to identify morphing attacks. MixFaceNet-MAD \cite{damer2022privacy} is a supervised MAD approach trained on the SMDD \cite{damer2022privacy} dataset that adapts the MixFaceNet \cite{boutros2021mixfacenets} backbone traditionally used for FR to the MAD task, through the usage of an extra binary classification layer. MADPromptS \cite{caldeira2025madprompts} takes advantage of CLIP's \cite{radford2021learning} zero-shot learning capabilities through efficient prompt engineering and aggregation, achieving SOTA results in MAD without fine-tuning the foundation model. These three strategies were chosen because they represent complementary approaches within MAD, covering unsupervised anomaly detection, supervised learning, and foundation model-based methods. We use the officially released models of these three methods \cite{caldeira2025madprompts, damer2022privacy, fang2022unsupervised} to perform MAD evaluation. Before being processed by the morphing attack detectors, each sample was cropped using the bounding boxes' information extracted by RetinaFace \cite{DBLP:conf/cvpr/DengGVKZ20} with 5\% extension of the width and height to include the whole face, following \cite{damer2022privacy}. All crops were resized to $224 \times 224$ and normalized following the hyperparameter values originally used by each approach. %(SPL and MixFaceNet-MAD: $\mu=[0.5, 0.5, 0.5]$, $\sigma=[0.5, 0.5, 0.5]$; MADPromptS: $\mu=[0.48145466, 0.4578275, 0.40821073]$, $\sigma=[0.26862954, 0.26130258, 0.27577711]$).

While traditional MAD evaluation datasets contain samples belonging to the two possible classes (bona fide and morphing attack), the datasets analyzed in this work only contain bona fide samples. This setting is representative of the real-world scenario of EES, where face image data is acquired through live capture at the airport gates \textcolor{black}{or in private/semi-private environments for individuals with low mobility}. Hence, common MAD metrics defined in the ISO/IEC 30107-3 \cite{ISO301073} standard, such as the equal error rate (EER) and the Attack Presentation Classification Error Rate (APCER) at a fixed BPCER, cannot be evaluated in this work's context. Moreover, following the traditional evaluation protocols to determine the BPCER at fixed APCER also requires the presence of attack samples in the evaluation dataset to determine the APCER thresholds. Due to the absence of attack samples, the APCER thresholds have to be determined in advance using a different dataset \textcolor{black}{containing attack samples}. Hence, we start by evaluating each of the considered MAD techniques on the complete MAD22 \cite{huber2022syn} dataset, including its original subsets (FaceMorpher \cite{quek2019facemorpher}, MIPGAN\_I \cite{DBLP:journals/tbbis/ZhangVRRDB21}, MIPGAN\_II \cite{DBLP:journals/tbbis/ZhangVRRDB21}, OpenCV \cite{openCVmorph}, and Webmorph \cite{debruine2018debruine}) and its extension MorDIFF \cite{damer2023mordiff}, to extract the thresholds that result in a maximum APCER of 1\%, 10\%, and 20\%. Since MixFaceNet-MAD and MADPromptS normalize the MAD scores before extracting the metrics, the mean and standard deviation values used in this process are also extracted. \textcolor{black}{Hence, the considered thresholds are determined based on information extracted from both bona fide and attack samples, while following the methodology used in real-world MAD application settings, where the classification thresholds and normalization statistics are pre-defined based on the evaluation of labeled datasets and subsequently applied in the field}. Then, each MAD method is used to determine the BPCER achieved by the original and segmented datasets by applying each of the extracted thresholds as a decision boundary between bona fide and morphing samples, following normalization with the extracted mean and standard deviation values, for MixFaceNet-MAD and MADPromptS. These metrics are referred to as BPCER at a fixed threshold (BPCER@t1\%, for example), from now on.

\begin{figure*}
    \centering
    \includegraphics[width=0.855\linewidth]{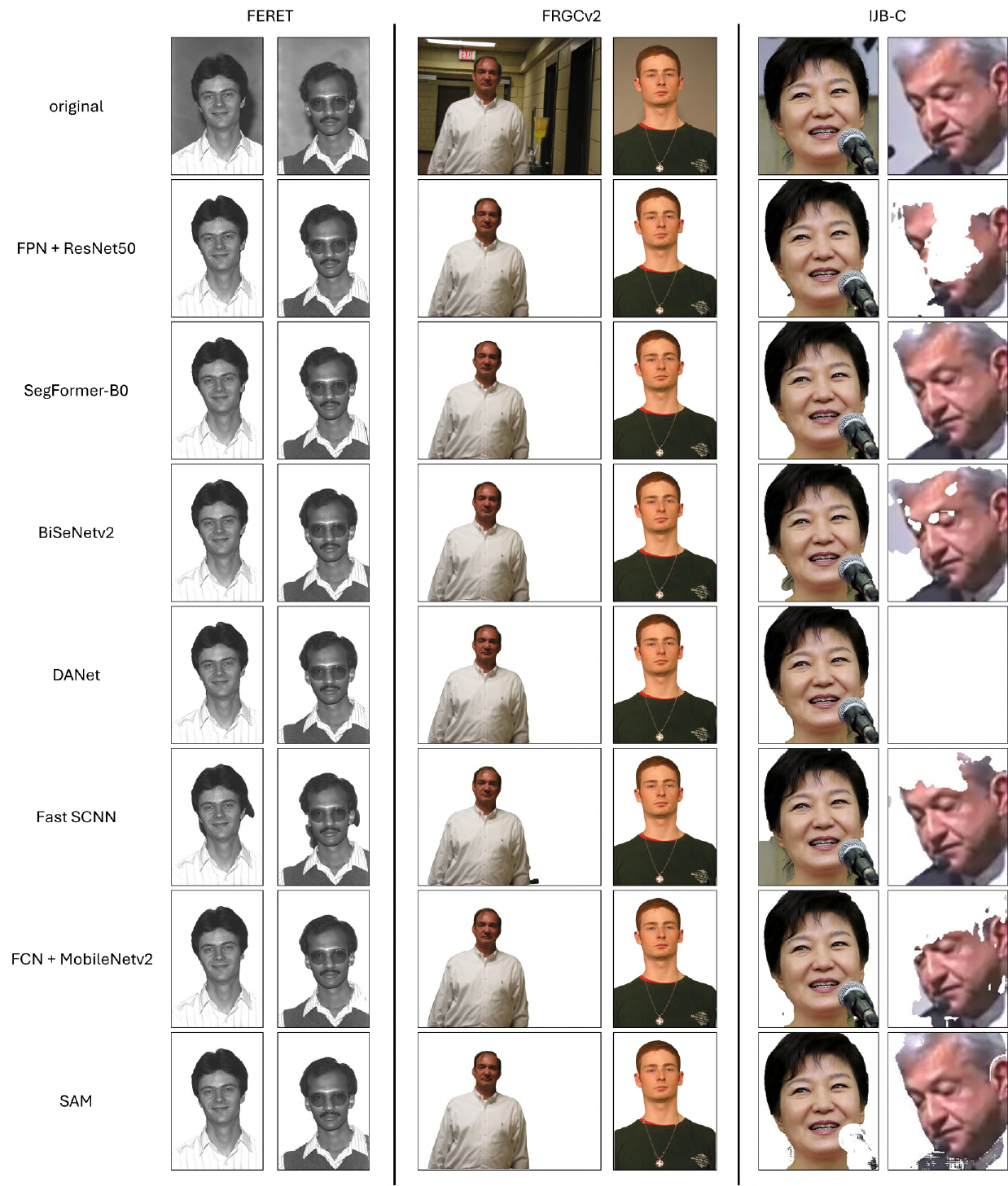}
    \vspace{-2mm}
    \caption{Pictures from the original FERET, FRGCv2 and IJB-C datasets before and after background removal. The margins were added to highlight the images' borders after background removal. Notice that the more controlled conditions on which FERET and FRGCv2 images were captured facilitate the segmentation task, resulting in efficient background removal by most techniques, and small face image quality and FR performance oscillations (Table \ref{tab:FR1}). For IJB-C, low-quality pictures (last column) are not easily segmented by all methods, with FPN + Resnet50, DANet, and FCN + MobileNetv2 resulting in significant facial occlusions. This phenomenon directly impacts average image quality and FR performance (Table \ref{tab:FR2}).}
    \vspace{-6mm}
    \label{fig:examples}
\end{figure*}

\begin{table}[t]
    \centering
    \tiny
    \caption{FR performance and FIQA metrics on FERET, FRGCv2
    and their segmented variants. %Each segmentation technique was used for the background removal task to substitute the original image background with white pixels. 
    It can be seen that these metrics are not significantly impacted by background removal. %process.%, which generates images with an average quality similar to the original datasets. %These results are complemented by the more discriminative analysis presented in Table \ref{tab:gen_imp}.
    }
    \vspace{-2mm}
    \resizebox{0.99\linewidth}{!}{%
    \begin{tabular}{cc|cccc|cc} % Adjusted to 6 columns to match your header
    \hline
    \multirow{2}{*}{\textbf{Dataset}} & 
    \multirow{2}{*}{\textbf{Segmentation Method}} & \multicolumn{4}{|c}{\textbf{FR Performance (\%)}} & \multicolumn{2}{|c}{\textbf{FIQA}} \\ 
    & & \textbf{ElasticFace} & \textbf{ArcFace} & \textbf{SwinFace} & \textbf{TransFace} & \textbf{avg} & \textbf{std}\\ \hline 
    
    \multirow{8}{*}{FERET} & 
    -- & 100.00 & 100.00 & 100.00 & 100.00 & 2.370 & 0.074 \\ \hline %\cline{2-8}
     &FPN + ResNet50 \cite{lin2017feature} & 100.00 & 100.00 & 100.00 & 100.00 & 2.369 & 0.074 \\
     &SegFormer-B0 \cite{xie2021segformer} & 100.00 & 100.00 & 100.00 & 100.00 & 2.369 & 0.074 \\ 
     &BiSeNetv2 \cite{yu2018bisenet} & 100.00 & 100.00 & 100.00 & 100.00 & 2.367 & 0.075\\ 
     &DANet \cite{fu2019dual} & 100.00 & 100.00 & 100.00 & 100.00 & 2.369 & 0.074 \\
     &Fast SCNN \cite{DBLP:conf/bmvc/PoudelLC19} & 100.00 & 100.00 & 100.00 & 100.00 & 2.368 & 0.074 \\ 
     &FCN + MobileNetv2 \cite{long2015fully} & 100.00 & 100.00 & 100.00 & 100.00 & 2.368 & 0.074 \\
     &SAM \cite{kirillov2023segment} & 100.00 & 100.00 & 100.00 & 100.00 & 2.362 & 0.076 \\ \hline \hline
    \multirow{8}{*}{FRGCv2} & -- & 99.98 & 99.98 &	99.98 &	99.98 & 2.329 & 0.071 \\ \cline{2-8}
    & FPN + ResNet50 \cite{lin2017feature} & 99.99 & 99.99 & 99.99 & 99.99 &  2.329 & 0.070\\
    & SegFormer-B0 \cite{xie2021segformer}  & 99.99 & 99.99 & 99.99 & 100.00 & 2.329 & 0.070\\ 
    & BiSeNetv2 \cite{yu2018bisenet} & 99.99 & 99.99 & 99.99 & 99.99 & 2.329 & 0.071 \\ 
    & DANet \cite{fu2019dual} & 99.99 & 99.99 &	99.99 &	99.99 & 2.330 & 0.070\\
    & Fast SCNN \cite{DBLP:conf/bmvc/PoudelLC19} & 99.99 & 99.99 & 99.99 &	99.99 & 2.328 & 0.071\\ 
    & FCN + MobileNetv2 \cite{long2015fully} & 99.99 & 99.99 &	99.99 & 100.00 & 2.329 & 0.070\\
    & SAM \cite{kirillov2023segment} & 99.98 & 99.98 & 99.98 & 99.98 & 2.330 & 0.073\\ \hline
    \end{tabular}}
    \label{tab:FR1}
    \vspace{-5mm}
\end{table}

\section{Results}
\label{sec:results}

\subsection{Face Recognition}

\textbf{Controlled Scenario:} The FR performance and FIQA results of each analyzed network when evaluating FERET %FRGCv2, and their 
and its segmented versions are presented in Table \ref{tab:FR1}. It can be seen that no error is detected when evaluating the FERET dataset's pairs with the considered FR approaches, regardless of the segmentation technique used for background removal. These results are aligned with the FIQA statistics of FERET's segmented versions, which present marginal variations in comparison with the original dataset. While revealing the low impact that background removal has on the FR system's capacity to correctly evaluate FERET, these results are consistent with the controlled conditions under which FERET's images were acquired (low background noise, frontal position), which make it an ideal dataset from the perspective of both the FR system and the segmentation network. This is visually supported by Figure \ref{fig:examples}, where the majority of the considered segmentation techniques manage to efficiently segment the background, resulting in passport-like pictures. %Regarding FRGC, it can be seen that ...
As mentioned in Section \ref{sec:methodology}, this analysis can be complemented by reporting the average cosine similarity for genuine and impostor pairs, and the difference between them ($\Delta$). While the results presented in Table \ref{tab:FR1} show that no background removal technique results in a FR performance degradation, it can be argued that they are based on a low amount of samples (2,722 images) whose characteristics perfectly depict the desired features that the images should have to facilitate the background removal task. Small deviations might arise from the background removal steps and be invisible in the FR performance evaluation if they are not enough to generate an overlapping region between genuine and impostor distributions. If these curves move towards each other after segmentation (smaller $\Delta$), the background removal task resulted in a less discriminative FR capacity, which might not directly translate into changes in FR verification accuracy. As seen in Table \ref{tab:gen_imp}, the average genuine cosine similarity decreases after each of the segmentation techniques is applied. \textcolor{black}{This tendency is generally accompanied by a slight increase in average impostor scores, resulting in a decrease in $\Delta$ in all the evaluated scenarios}. These results expose a slight approximation tendency between the genuine and impostor curves after background removal, highlighting the existence of a negative impact on the FR scores distribution when segmentation is performed.

\begin{table*}[t]
    \centering
    \tiny
    \caption{\textcolor{black}{$gen_{avg}$, $imp_{avg}$ and $\Delta$ on FERET, FRGCv2, IJB-C, and their segmented variants. The values inside the parentheses represent the variation in magnitude of each metric compared to the baseline (no segmentation). While Table \ref{tab:FR1} does not reveal any impact on performance associated with the background removal process on FERET and FRGCv2, these results allow for a more detailed analysis, as they highlight the approximation tendency between the mean genuine and impostor pairs' cosine similarity values, revealing the lower discriminative capacity of the FR systems when evaluating segmented data. For IJB-C, the magnitude of this approximation is directly correlated with the verified FR performance drops (Table \ref{tab:FR2}).}}
    \vspace{-2mm}
    \resizebox{0.99\linewidth}{!}{%
    \begin{tabular}{cc|ccc|ccc|ccc|ccc} 
    \hline
    \multirow{2}{*}{\textbf{Dataset}} & \multirow{2}{*}{\textbf{Segmentation Method}} & \multicolumn{3}{|c}{\textbf{ElasticFace}} & \multicolumn{3}{|c}{\textbf{ArcFace}} & \multicolumn{3}{|c}{\textbf{SwinFace}} & \multicolumn{3}{|c}{\textbf{TransFace}}\\ 
     & & $gen_{avg}$ & $imp_{avg}$ & $\Delta$ & $gen_{avg}$ & $imp_{avg}$ & $\Delta$ & $gen_{avg}$ & $imp_{avg}$ & $\Delta$ & $gen_{avg}$ & $imp_{avg}$ & $\Delta$ \\ \hline
    \multirow{8}{*}{FERET} & -- & 0.8565 & 0.0110 & 0.8455 & 0.8784 & 0.0096 & 0.8688 & 0.8816 & 0.0145 & 0.8671 & 0.8650 & 0.0144 & 0.8506\\ \cline{2-14}
    & FPN + ResNet50 \cite{lin2017feature} & 0.8560 (-0.0005) & 0.0111 (+0.0001) & 0.8449 (-0.0006) & 0.8783 (-0.0001) & 0.0100 (+0.0004) & 0.8683 (-0.0005) & 0.8807 (-0.0009) & 0.0144 (-0.0001) & 0.8663 (-0.0008) & 0.8642 (-0.0008) & 0.0154 (+0.0010)& 0.8488 (-0.0018)\\
    & SegFormer-B0 \cite{xie2021segformer} & 0.8557 (-0.0008) & 0.0112 (+0.0002) & 0.8445 (-0.0010) & 0.8780 (-0.0004) & 0.0099 (+0.0003)& 0.8681 (-0.0007) & 0.8806 (-0.0010) & 0.0145 & 0.8661 (-0.0010) & 0.8643 (-0.0007) & 0.0154 (+0.0010) & 0.8489 (-0.0017)\\ 
    & BiSeNetv2 \cite{yu2018bisenet} & 0.8544 (-0.0021) & 0.0111 (+0.0001) & 0.8433 (-0.0022) &0.8766 (-0.0018) & 0.0098 (+0.0002) & 0.8668 (-0.0020) & 0.8796 (-0.0020) & 0.0144 (-0.0001) & 0.8652 (-0.0019) & 0.8631 (-0.0019) & 0.0153 (+0.0009) & 0.8478 (-0.0028) \\ 
    & DANet \cite{fu2019dual} & 0.8560 (-0.0005) & 0.0111 (+0.0001) & 0.8449 (-0.0006) & 0.8782 (-0.0002) & 0.0099 (+0.0003) & 0.8683 (-0.0005) & 0.8808 (-0.0008) & 0.0144 (-0.0001) & 0.8664 (-0.0007) & 0.8644 (-0.0006) & 0.0153 (+0.0009) & 0.8491 (-0.0015) \\
    & Fast SCNN \cite{DBLP:conf/bmvc/PoudelLC19} & 0.8555 (-0.0010) & 0.0111 (+0.0001) & 0.8444 (-0.0011) &0.8776 (-0.0008) & 0.0099 (+0.0003) & 0.8677 (-0.0011) & 0.8805 (-0.0011) & 0.0145 & 0.8660 (-0.0011) & 0.8639 (-0.0011) & 0.0153 (+0.0009) & 0.8486 (-0.0020) \\ 
    & FCN + MobileNetv2 \cite{long2015fully} & 0.8556 (-0.0009) & 0.0112 (+0.0002) & 0.8444 (-0.0011) & 0.8778 (-0.0006) & 0.0099 (+0.0003) & 0.8679 (-0.0009) & 0.8805 (-0.0011) & 0.0144 (-0.0001) & 0.8661 (-0.0010) & 0.8642 (-0.0008) & 0.0153 (+0.0009) & 0.8489 (-0.0017)\\
    & SAM \cite{kirillov2023segment} & 0.8498 (-0.0067) & 0.0110 & 0.8388 (-0.0067) & 0.8699 (-0.0085) & 0.0100 (+0.0004) & 0.8599 (-0.0089) & 0.8755 (-0.0061) & 0.0137 (-0.0008) & 0.8618 (-0.0053) & 0.8539 (-0.0011) & 0.0147 (+0.0003) & 0.8392 (-0.0014)\\ \hline \hline

     %FRGC
     
     \multirow{8}{*}{FRGCv2} & -- &  0.7402 & 0.0211	& 0.7191 & 0.7841 &	0.0204 & 0.7637 & 0.7825 & 0.0272 & 0.7553 & 0.7672 & 0.0254  & 0.7418 \\ \cline{2-14}
     & FPN + ResNet50 \cite{lin2017feature} &  0.7393 (-0.0009) & 0.0228 (+0.0017) & 0.7165 (-0.0026) & 0.7840 (-0.0001) & 0.0222 (+0.0018) & 0.7618 (-0.0019) & 0.7822 (-0.0003) & 0.0267 (-0.0005) & 0.7555 (+0.0002) & 0.7661 (-0.0011) & 0.0255 (+0.0001) & 0.7406 (-0.0012) \\
     & SegFormer-B0 \cite{xie2021segformer} &  0.7390 (-0.0012) & 0.0237 (+0.0026) & 0.7153 (-0.0038) & 0.7838 (-0.0003) & 0.0225 (+0.0021) & 0.7613 (-0.0024) & 0.7819 (-0.0006) & 0.0270 (-0.0002) & 0.7549 (-0.0004) & 0.7658 (-0.0014) & 0.0258 (+0.0004) & 0.7400 (-0.0018) \\ 
     & BiSeNetv2 \cite{yu2018bisenet} & 0.7393 (-0.0009) & 0.0238 (+0.0027)  & 0.7155 (-0.0036) & 0.7839 (-0.0002) & 0.0221 (+0.0017) & 0.7618 (-0.0019) & 0.7819 (-0.0006) & 0.0266 (-0.0006) & 0.7553 & 0.7661 (-0.0011) & 0.0261 (+0.0007) & 0.7400 (-0.0018) \\ 
     & DANet \cite{fu2019dual} & 0.7395 (-0.0007) & 0.0227 (+0.0016) & 0.7168 (-0.0023) & 0.7841 & 0.0217 (+0.0013) & 0.7624 (-0.0013) & 0.7822 (-0.0003)	& 0.0261 (-0.0011) & 0.7561 (+0.0008) & 0.7665 (-0.0007) & 0.0257 (+0.0003) & 0.7408 (-0.0010) \\
     & Fast SCNN \cite{DBLP:conf/bmvc/PoudelLC19} & 0.7388 (-0.0014)& 0.0241 (+0.0030) & 0.7147 (-0.0044)& 0.7835 (-0.0006) & 0.0221 (+0.0017) & 0.7614 (-0.0023) & 0.7816 (-0.0009) & 0.0267 (-0.0005)  & 0.7549 (-0.0004) & 0.7656 (-0.0016)	& 0.0261 (+0.0007) & 0.7395 (-0.0023) \\ 
     & FCN + MobileNetv2 \cite{long2015fully} &  0.7389 (-0.0013) & 0.0239 (+0.0028) & 0.7150 (-0.0041) & 0.7838 (-0.0003) & 0.0222 (+0.0018) & 0.7616 (-0.0021) & 0.7817 (-0.0008) & 0.0266 (-0.0006) & 0.7551 (-0.0002) & 0.7657 (-0.0015) & 0.0261 (+0.0007) & 0.7396 (-0.0022)\\
     & SAM \cite{kirillov2023segment} & 0.7383 (-0.0019) & 0.0231 (+0.0020) & 0.7152 (-0.0039) & 0.7830 (-0.0011) & 0.0233 (+0.0029) & 0.7597 (-0.0040) & 0.7817 (-0.0008) & 0.0265 (-0.0007) & 0.7552 (-0.0001) & 0.7633 (-0.0039) & 0.0251 (-0.0003) & 0.7382 (-0.0036) \\ \hline \hline

     %IJBC
     
    \multirow{8}{*}{IJB-C} & -- &  0.7160 & 0.0032  & 0.7128 & 0.7576	& 0.0028 & 0.7548 & 0.7476 & 0.0034 & 0.7442 & 0.7365 & 0.0024 & 0.7341 \\ \cline{2-14}
    & FPN + ResNet50 \cite{lin2017feature} & 0.6855 (-0.0305) & 0.0077 (+0.0045) & 0.6778 (-0.0350) &  0.7267 (-0.0309) & 0.0080 (+0.0052) & 0.7187 (-0.0361) &  0.7145  (-0.0331) &  0.0069 (+0.0035) & 0.7076 (-0.0366) &  0.6941 (-0.0424) &  0.0095 (+0.0071) & 0.6846 (-0.0495) \\
    & SegFormer-B0 \cite{xie2021segformer} & 0.7054 (-0.0106) & 0.0045 (+0.0013) & 0.7009 (-0.0119) &  0.7469 (-0.0107) & 0.0038 (+0.0010) & 0.7431 (-0.0117) &  0.7378  (-0.0098) & 0.0043  (+0.0009) & 0.7335 (-0.0107) &  0.7217 (-0.0148) &  0.0035 (+0.0011) & 0.7182 (-0.0159) \\ 
    & BiSeNetv2 \cite{yu2018bisenet} & 0.6858 (-0.0302) & 0.0063 (+0.0031) & 0.6795 (-0.0333) & 0.7289 (-0.0287) & 0.0051 (+0.0023) & 0.7238 (-0.0310) &   0.7190 (-0.0286) & 0.0052 (+0.0018) & 0.7138 (-0.0304) &  0.7036 (-0.0329) & 0.0047  (+0.0023) & 0.6989 (-0.0352) \\ 
    & DANet \cite{fu2019dual} & 0.5696 (-0.1464) & 0.0722 (+0.0690) & 0.4974 (-0.2154) & 0.6145 (-0.1431) & 0.0835 (+0.0807) & 0.5310 (-0.2238) &  0.5807  (-0.1669) &  0.0478 (+0.0444) & 0.5329 (-0.2113) &   0.5851 (-0.1514) &  0.1013 (+0.0989) & 0.4838 (-0.2503) \\
    & Fast SCNN \cite{DBLP:conf/bmvc/PoudelLC19} & 0.6929 (-0.0231) & 0.0066 (+0.0034) & 0.6863 (-0.0265) & 0.7351 (-0.0225) & 0.0052 (+0.0024) & 0.7299 (-0.0249) &  0.7257  (-0.0219) &  0.0053 (+0.0019) & 0.7204 (-0.0238) &  0.7097 (-0.0268) &  0.0044 (+0.0020) & 0.7053 (-0.0288) \\ 
    & FCN + MobileNetv2 \cite{long2015fully} & 0.6144 (-0.1016) & 0.0255 (+0.0223) & 0.5889 (-0.1239) & 0.6606 (-0.0970) & 0.0301 (+0.0273) & 0.6305 (-0.1243) & 0.6387 (-0.1089) & 0.0157 (+0.0123) & 0.6230 (-0.1212) &  0.6223 (-0.1142) &  0.0280 (+0.0256) & 0.5943 (-0.1398) \\
    & SAM \cite{kirillov2023segment} & 0.7082 (-0.0078) & 0.0042 (+0.0010) & 0.7040 (-0.0088) & 0.7497 (-0.0079) & 0.0037 (+0.0009) & 0.7460 (-0.0088) &  0.7403  (-0.0073) &  0.0043 (+0.0009) & 0.7360 (-0.0082) & 0.7241  (-0.0124) &  0.0039 (+0.0015) & 0.7202 (-0.0139) \\ \hline 
    \end{tabular} }
    \vspace{-3mm}
    \label{tab:gen_imp}
\end{table*}

\begin{table}[t]
    \centering
    \caption{FR performance and FIQA metrics on IJB-C, and its segmented variants. %Each segmentation technique was used for the background removal task to substitute the original image background with white pixels. 
    The challenging IJB-C dataset results in significant differences in FR performance and face image quality, compared to FERET and FRGCv2 (Table \ref{tab:FR1}). SegFormer-B0 and SAM achieve the closest performance to the unsegmented data, while FPN + Resnet50, DANet, and FCN + MobileNetv2 result in extreme performance drops.}
    \vspace{-2mm}
    \resizebox{0.99\linewidth}{!}{%
    \begin{tabular}{c|cc|cc|cc|cc|cc} % Adjusted to 6 columns to match your header
    \hline
     \multirow{3}{*}{\textbf{Segmentation Method}} & \multicolumn{8}{|c|}{\textbf{FR Performance (\%)}}  & \\ 
    & \multicolumn{2}{c|}{\textbf{ElasticFace}} & \multicolumn{2}{c|}{\textbf{ArcFace}} & \multicolumn{2}{c|}{\textbf{SwinFace}} & \multicolumn{2}{c|}{\textbf{TransFace}} & \multicolumn{2}{|c}{\textbf{FIQA}} \\ 
    & $10^{-5}$ & $10^{-4}$ & $10^{-5}$ & $10^{-4}$ & $10^{-5}$ & $10^{-4}$ & $10^{-5}$ & $10^{-4}$ &  \textbf{avg} & \textbf{std} \\ \hline 
    -- & 94.67 & 96.47 & 95.25 & 96.71 & 94.86 & 96.71 & 96.38 & 97.61 & 2.025 & 0.395 \\ \hline
    FPN + ResNet50 \cite{lin2017feature} & 3.94 & 66.89 & 0.09 & 50.47 & 44.48 & 85.21 & 1.08 & 45.98 & 1.903 & 0.557 \\ 
    SegFormer-B0 \cite{xie2021segformer} & 93.04 & 95.82 & 94.02 & 96.10 & 94.16	& 96.11	& 94.99	& 97.04 & 2.003	& 0.418\\ 
    BiSeNetv2 \cite{yu2018bisenet} & 82.97 & 93.44 & 78.99 & 94.11 & 90.99 & 94.67 & 75.91 & 94.00 & 1.953 & 0.468\\ 
    DANet \cite{fu2019dual} & 0.01 & 0.08 & 0.01 & 0.05 & 0.20 & 3.37 & 0.01 & 0.08 & 1.399	& 0.712\\
    Fast SCNN \cite{DBLP:conf/bmvc/PoudelLC19} & 88.66 & 94.69 & 90.16 & 95.37 & 92.84 & 95.40 & 91.29 & 96.37 & 1.967 & 0.441 \\ 
    FCN + MobileNetv2 \cite{long2015fully} & 0.03	& 7.64 & 0.01 & 0.37 & 11.85	& 45.46	& 0.03 & 1.41 & 1.731 & 0.601\\
    SAM \cite{kirillov2023segment} & 93.90 & 96.23	& 94.64	& 96.49	& 94.55	& 96.50 & 95.74	& 97.27	& 2.014 & 0.397 \\ \hline 
    \end{tabular}}
    \vspace{-5mm}
    \label{tab:FR2}
\end{table}

\begin{table*}[t]
    \centering
    \tiny
    \caption{MAD triggering results of three MAD methods (SPL, MixFaceNet-MAD, and MADPromptS) on FERET, FRGCv2, IJB-C, and their segmented variants.Notice that background removal results in lower triggering rates of SPL and MixFaceNet-MAD by bona fide samples (lower BPCER at a fixed threshold). An opposite tendency is verified when performing the detection with MADPromptS. (*) For MixFaceNet-MAD, the only threshold that allowed to obtain an APCER of 1\% was positive infinity, resulting in BPCER@t1\% values of 100.00\% regardless of the evaluated dataset; hence, this metric does not allow to differentiate between the different strategies, and should be disregarded in the results' analysis of MixFaceNet-MAD.}
    \vspace{-2mm}
    \begin{tabular}{cc|ccc|ccc|ccc}
    \hline
    \multirow{3}{*}{\textbf{Dataset}} & \multirow{3}{*}{\textbf{Segmentation Method}} & \multicolumn{9}{|c}{\textbf{MAD Performance (\%)}} \\ 
    & & \multicolumn{3}{c|}{\textbf{SPL}} & \multicolumn{3}{c|}{\textbf{MixFaceNet-MAD}} & \multicolumn{3}{c}{\textbf{MADPromptS}} \\ 
    & & BPCER@t1\% & BPCER@t10\% & BPCER@t20\% & BPCER@t1\%* & BPCER@t10\% & BPCER@t20\% & BPCER@t1\% & BPCER@t10\% & BPCER@t20\% \\
    \hline 
    
    \multirow{8}{*}{FERET} & -- & 45.89	& 25.13 & 18.07	& 100.00 & 18.88 & 10.10 & 36.22 & 4.81	& 1.18 \\ \cline{2-11}
    & FPN + ResNet50 \cite{lin2017feature} & 29.50 & 16.72 & 13.15 & 100.00 & 11.98 & 6.54 & 55.69 & 8.67 & 2.57\\
    & SegFormer-B0 \cite{xie2021segformer} & 28.91 & 16.79 & 13.30 & 100.00 & 12.01 & 6.25 & 56.43 & 8.34 & 2.50\\ 
    & BiSeNetv2 \cite{yu2018bisenet} & 25.72	& 15.69	& 12.71	& 100.00 & 11.46 & 6.76 & 54.70 & 8.41 & 2.28\\ 
    & DANet \cite{fu2019dual} & 25.09	& 15.94	& 13.04	& 100.00 & 11.65 & 6.58 & 54.52	& 7.79 & 2.39\\
    & Fast SCNN \cite{DBLP:conf/bmvc/PoudelLC19} & 26.49	& 15.47	& 12.49	& 100.00 & 11.09 & 6.32 & 53.38 & 7.38 & 2.17\\ 
    & FCN + MobileNetv2 \cite{long2015fully} & 25.20	& 15.28	& 12.09	& 100.00 & 10.43 & 5.69 & 54.59	& 8.01 & 2.79\\
    & SAM \cite{kirillov2023segment} & 18.81 & 9.70 &	7.83 & 100.00 &	7.27 & 3.89 & 62.05	& 12.27	& 5.18 \\ \hline
    
    \multirow{8}{*}{FRGCv2} & -- & 65.19 & 45.87 & 38.22 & 100.00 & 49.80 & 31.96 & 36.95 & 1.21 & 0.17\\ \cline{2-11}
    & FPN + ResNet50 \cite{lin2017feature} & 28.52 &  17.74	& 13.47	& 100.00 & 6.25	& 2.81 & 70.98 & 11.03 & 2.94\\
    &  SegFormer-B0 \cite{xie2021segformer} & 27.34	& 16.56	& 12.62	& 100.00 & 5.51	& 2.34 & 71.78 & 11.36 & 3.10\\ 
    & BiSeNetv2 \cite{yu2018bisenet} & 27.57 & 16.20 & 12.00 & 100.00 &	5.79 & 2.71 & 70.74	& 10.42	& 2.76 \\ 
    & DANet \cite{fu2019dual} & 27.70 &	16.32 &	12.24 &	100.00	& 6.13	& 2.82 & 69.48 & 9.73 & 2.49\\
    & Fast SCNN \cite{DBLP:conf/bmvc/PoudelLC19} & 27.20 &	16.17 & 12.12 & 100.00 & 6.01 & 2.77 & 71.75 & 11.02 & 2.88\\ 
    & FCN + MobileNetv2 \cite{long2015fully} & 27.07 & 15.69 & 11.65 & 100.00 & 5.71 & 2.64 & 71.53 & 10.90 & 2.85 \\
    & SAM \cite{kirillov2023segment} & 18.03 & 10.92 & 8.03 & 100.00 &	3.33 &  1.34 & 76.00 &	22.83 & 11.89\\ \hline
    
     \multirow{8}{*}{IJB-C} & -- & 86.98 & 81.83 & 78.73 & 100.00 & 74.29	& 63.03 & 7.82 & 0.53 & 0.17 \\ \cline{2-11}
    & FPN + ResNet50 \cite{lin2017feature} & 24.62 & 17.07 & 14.13 & 100.00 & 14.39 & 5.97 & 50.34 & 14.05 & 7.13 \\
    & SegFormer-B0 \cite{xie2021segformer} & 33.83 & 23.93 & 19.96 & 100.00	& 21.71	& 12.93 & 42.23	& 8.02 & 3.50\\ 
    & BiSeNetv2 \cite{yu2018bisenet} & 30.10 & 21.72 & 18.25 & 100.00 & 21.49 & 12.90 & 46.26 & 11.70 & 5.91\\ 
    & DANet \cite{fu2019dual} & 21.72 & 18.96 & 17.89 & 100.00 & 10.56 & 0.58 & 60.00 & 25.91 & 14.98\\
    & Fast SCNN \cite{DBLP:conf/bmvc/PoudelLC19} & 26.60 & 18.36 & 15.03 & 100.00 &	17.48 & 9.79 & 51.15 & 12.42 & 5.94\\ 
    & FCN + MobileNetv2 \cite{long2015fully} & 13.02 & 9.01 & 7.58 & 100.00	& 5.31 & 1.28 & 54.59 & 8.01 & 2.79\\
    & SAM \cite{kirillov2023segment} & 21.01 & 14.01 & 11.44 & 100.00 & 14.09 & 8.39 & 53.03 & 10.23 & 4.15 \\ \hline
    \end{tabular}
    \vspace{-5mm}
    \label{tab:MAD}
\end{table*}

\textbf{Semi-Controlled Scenario:} \textcolor{black}{Table \ref{tab:FR1} also summarizes the FR performance and quality metrics results on FRGCv2. Similarly to FERET, it can be seen that the quality of the obtained samples does not fluctuate significantly after segmentation. This tendency is also associated with negligible fluctuations in FR performance. Hence, it is possible to conclude that while the FRGCv2 dataset contains pictures with an unconstrained background, its semi-controlled environment (high-quality samples with frontal pose) allows segmentation techniques to work effectively without compromising the images' quality, resulting in minimal effects in downstream FR performance. Nonetheless, the results depicted in Table \ref{tab:gen_imp} highlight a slight approximation tendency between the average genuine and impostor scores (smaller $\Delta$) for the majority of the evaluated scenarios, revealing the negative impact of segmentation on the FR scores distribution.} 

\textbf{Uncontrolled Scenario:} Table \ref{tab:FR2} summarizes the evaluation results of the considered FR models on the IJB-C dataset. While all the considered segmentation strategies resulted in drops in FR performance, the extent of the decrease in performance varies significantly between segmentation networks. Datasets segmented with SegFormer-B0 and SAM achieve the closest performance to the unsegmented data, although with non-negligible performance drops. The superiority of these methods in maintaining FR performance integrity is supported by the higher average FIQA scores achieved by these strategies in comparison to the remaining methods. BiSeNetv2 and Fast SCNN are associated with a further drop in FR performance and quality, while FPN + Resnet50, DANet, and FCN + MobileNetv2 result in a complete lack of recognition capacity. \textcolor{black}{These results are accompanied by a significant decrease in the value of $\Delta$ for all the analyzed techniques, which directly translates to a higher overlap between genuine and impostor scores distributions, as visually supported in the supplementary material. 
The decrease in $\Delta$ is particularly high for DANet, FCN + MobileNetv2, and FPN + Resnet50, which aligns with the poor FR performance on images segmented with these techniques.} These results are also visually supported by Figure \ref{fig:examples}, which displays some segmentation examples of IJB-C images using each of the seven considered methods. While images segmented by SegFormer-B0 and SAM are visually consistent, low-quality pictures, such as the one presented in the last column, are very poorly segmented by FPN + Resnet50, DANet, and FCN + MobileNetv2, resulting in significant facial occlusions.

These results show that when faced with more challenging images, such as those contained in IJB-C, background removal techniques exhibit a visible negative impact on FR performance and image quality. The differences in the results between the analyzed segmentation techniques highlight that the choice of an appropriate segmentation method is essential to mitigate image quality deterioration and, thus, FR performance drops. While for datasets acquired in extremely controlled conditions, such as FERET, the effects of background removal do not significantly affect the obtained results, images acquired under more challenging conditions expose the importance of properly assessing the segmentation capacity of these systems in uncontrolled environments before their deployment in highly sensitive applications such as ESS for border control. The degradation in FR performance after background removal may be attributed to the limited accuracy of the segmentation models. In particular, models such as DANet fail to reliably separate the background in challenging samples, as illustrated in Figure \ref{fig:examples}. A complementary evaluation of the segmentation performance of all segmentation networks used in this work can be found in the supplementary material.

\subsection{Morphing Attack Detection Triggering}

Table \ref{tab:MAD} shows how the trigger frequency of MAD systems changes when performing background removal. %with different strategies. 
While the FR performance results differed significantly between the three studied datasets, they present a similar tendency regarding MAD systems' triggering. %Overall, it can be seen that 
SPL and MixFaceNet-MAD benefit from the usage of background removal techniques, as they are significantly less triggered to incorrectly detect morphing attacks when segmentation is performed, regardless of the used segmentation technique. In particular, SAM works the best for FERET, resulting in a decrease of 15.43 and 11.61 percentage points in terms of BPCER@t10\% for SPL and MixFaceNet-MAD, respectively. \textcolor{black}{A similar tendency is verified for FRGCv2.} This is also the case for the IJB-C benchmark if we exclude the results of FCN + MobileNetv2 and DANet, due to their extremely reduced FR performance. MADPromptS, however, is negatively affected by background removal. This impact is particularly evident when analyzing the BPCER@t1\%, which rises from 7.82\% when no segmentation is performed on IJB-C to 42.23\% when segmenting with SegFormer-B0, and even higher values for %when using any of 
the remaining methods. These results suggest that SPL and MixFaceNet-MAD tend to get triggered by background information that might be erroneously interpreted as containing morphing artifacts. When this information is substituted by a white background, these systems stop being triggered by such cues, resulting in significantly lower BPCER values. MADPromptS works in a fundamentally different way, as it relies on CLIP's built-in knowledge to classify each sample. Hence, the obtained results show that the pre-trained ViT large architecture of CLIP tends to associate white backgrounds with morphing cues, resulting in a significant MAD triggering increase for segmented samples. 

Two main conclusions can be drawn from the obtained results. On one side, it is essential to test the MAD systems applied on top of border control images on segmented datasets before their deployment in EES applications, since different MAD systems present significantly different behaviors when evaluating the same segmented data, and the background removal can prove beneficial or impairing depending on the evaluated system. On the other side, the benefit brought by the background removal step to SPL and MixFaceNet-MAD is far from bringing these methods to the same performance level as MADPromptS on the original data. In fact, FERET and FRGCv2 segmented with SAM and evaluated by SPL or MixFaceNet-MAD only outperform the MADPromptS evaluation of the original data in one out of six metrics. For IJB-C, MADPromptS applied to the original data is never surpassed by SPL or MixFaceNet-MAD, regardless of the chosen segmentation technique. The combination of these conclusions shows that while some methods benefit from the background removal, they may still fall behind state-of-the-art MAD strategies applied to non-segmented data, which limits the applicability of such strategies in real use-cases.
\section{Conclusion}
\label{sec:conclusion}

Large-scale biometric systems rely on high-quality facial images to ensure reliable enrollment and recognition. To support this, standards such as ISO/IEC 19794-5 enforce strict requirements, including the use of plain, light-colored backgrounds. However, in practical deployment scenarios such as enrollment at airports or border control environments, these controlled capture conditions cannot always be guaranteed. This motivates the use of face segmentation-based background removal to generate standardized, compliant facial images from unconstrained captures. Yet, a critical question arises: does such preprocessing preserve both recognition performance and the robustness of security mechanisms, particularly against attacks such as face morphing?

We addressed this question through a comprehensive evaluation of multiple segmentation techniques across diverse datasets and biometric tasks. Our results demonstrate that background removal is not a neutral operation. While its impact is minimal under controlled conditions, it leads to measurable FR performance and image quality degradation in unconstrained scenarios, with strong dependence on the segmentation method. At the same time, its effect on MAD is highly method-dependent: some approaches benefit from reduced false triggering due to the removal of misleading background cues, while others, especially foundation-model-based methods, show significant performance degradation.

These findings highlight a fundamental trade-off between enforcing image appearance and preserving the reliability and security of biometric systems. Consequently, background removal should not be treated as a standalone preprocessing step, but rather as an integral component of the biometric pipeline that must be jointly evaluated with both recognition and attack detection systems. From an operational perspective, our work underscores the need for careful validation of such techniques before deployment in real-world systems such as border control. Future research should focus on developing segmentation and preprocessing strategies that are explicitly optimized for downstream biometric tasks, as well as designing recognition and security mechanisms that are robust to such transformations.

\section*{Ethical Impact Statement}

\textcolor{black}{This work provides a comprehensive analysis of the impact of background removal in downstream border control tasks, to assess the feasibility of applying segmentation-based background removal in real-world applications such as entry–exit systems. At the same time, we recognize that face recognition technologies in general may be susceptible to misuse. Such misuse may include the processing of biometric data beyond legal, consensual, and ethical boundaries for the deployment of functionalities that extend beyond legitimate identity verification \cite{meden2021privacy}. In this context, we emphasize our commitment to ethical standards and strongly reject any application that violates legal frameworks or fundamental rights.}

% Ensure egbib.bib exists in your Overleaf file list!
\bibliographystyle{ieee}
\bibliography{egbib}

@String(CVPR= {IEEE Conf. Comput. Vis. Pattern Recog.})

@String(ICCV= {Int. Conf. Comput. Vis.})

@String(ECCV= {Eur. Conf. Comput. Vis.})

@String(ICPR = {Int. Conf. Pattern Recog.})

@String(BMVC= {Brit. Mach. Vis. Conf.})

@String(ICASSP=	{ICASSP})

@String(ICLR = {Int. Conf. Learn. Represent.})

@String(CVPR  = {CVPR})

@String(ICCV  = {ICCV})

@String(ECCV  = {ECCV})

@String(ICPR  = {ICPR})

@String(BMVC  =	{BMVC})

@String(ICLR  = {ICLR})

@article{xie2021segformer,
  title={SegFormer: Simple and efficient design for semantic segmentation with transformers},
  author={Xie, Enze and Wang, Wenhai and Yu, Zhiding and Anandkumar, Anima and Alvarez, Jose M and Luo, Ping},
  journal={Advances in neural information processing systems},
  volume={34},
  pages={12077--12090},
  year={2021}
}

@inproceedings{yu2018bisenet,
  title={Bisenet: Bilateral segmentation network for real-time semantic segmentation},
  author={Yu, Changqian and Wang, Jingbo and Peng, Chao and Gao, Changxin and Yu, Gang and Sang, Nong},
  booktitle={Proceedings of the European conference on computer vision (ECCV)},
  pages={325--341},
  year={2018}
}

@inproceedings{lin2017feature,
  title={Feature pyramid networks for object detection},
  author={Lin, Tsung-Yi and Doll{\'a}r, Piotr and Girshick, Ross and He, Kaiming and Hariharan, Bharath and Belongie, Serge},
  booktitle={Proceedings of the IEEE conference on computer vision and pattern recognition},
  pages={2117--2125},
  year={2017}
}

@inproceedings{long2015fully,
  title={Fully convolutional networks for semantic segmentation},
  author={Long, Jonathan and Shelhamer, Evan and Darrell, Trevor},
  booktitle={Proceedings of the IEEE conference on computer vision and pattern recognition},
  pages={3431--3440},
  year={2015}
}

@inproceedings{fu2019dual,
  title={Dual attention network for scene segmentation},
  author={Fu, Jun and Liu, Jing and Tian, Haijie and Li, Yong and Bao, Yongjun and Fang, Zhiwei and Lu, Hanqing},
  booktitle={Proceedings of the IEEE/CVF conference on computer vision and pattern recognition},
  pages={3146--3154},
  year={2019}
}

@inproceedings{kirillov2023segment,
  title={Segment anything},
  author={Kirillov, Alexander and Mintun, Eric and Ravi, Nikhila and Mao, Hanzi and Rolland, Chloe and Gustafson, Laura and Xiao, Tete and Whitehead, Spencer and Berg, Alexander C and Lo, Wan-Yen and others},
  booktitle={Proceedings of the IEEE/CVF international conference on computer vision},
  pages={4015--4026},
  year={2023}
}

@inproceedings{DBLP:conf/bmvc/PoudelLC19,
  author       = {Rudra P. K. Poudel and
                  Stephan Liwicki and
                  Roberto Cipolla},
  title        = {Fast-SCNN: Fast Semantic Segmentation Network},
  booktitle    = {{BMVC}},
  pages        = {289},
  publisher    = {{BMVA} Press},
  year         = {2019}
}

@inproceedings{DBLP:conf/cvpr/BoutrosDKK22,
  author       = {Fadi Boutros and
                  Naser Damer and
                  Florian Kirchbuchner and
                  Arjan Kuijper},
  title        = {ElasticFace: Elastic Margin Loss for Deep Face Recognition},
  booktitle    = {{CVPR} Workshops},
  pages        = {1577--1586},
  publisher    = {{IEEE}},
  year         = {2022}
}

@inproceedings{DBLP:conf/cvpr/DengGXZ19,
  author       = {Jiankang Deng and
                  Jia Guo and
                  Niannan Xue and
                  Stefanos Zafeiriou},
  title        = {ArcFace: Additive Angular Margin Loss for Deep Face Recognition},
  booktitle    = {{CVPR}},
  pages        = {4690--4699},
  publisher    = {Computer Vision Foundation / {IEEE}},
  year         = {2019}
}

@article{DBLP:journals/tcsv/QinWDWCHD24,
  author       = {Lixiong Qin and
                  Mei Wang and
                  Chao Deng and
                  Ke Wang and
                  Xi Chen and
                  Jiani Hu and
                  Weihong Deng},
  title        = {SwinFace: {A} Multi-Task Transformer for Face Recognition, Expression
                  Recognition, Age Estimation and Attribute Estimation},
  journal      = {{IEEE} Trans. Circuits Syst. Video Technol.},
  volume       = {34},
  number       = {4},
  pages        = {2223--2234},
  year         = {2024}
}

@inproceedings{DBLP:conf/iccv/DanLXD0XS23,
  author       = {Jun Dan and
                  Yang Liu and
                  Haoyu Xie and
                  Jiankang Deng and
                  Haoran Xie and
                  Xuansong Xie and
                  Baigui Sun},
  title        = {TransFace: Calibrating Transformer Training for Face Recognition from
                  a Data-Centric Perspective},
  booktitle    = {{ICCV}},
  pages        = {20585--20596},
  publisher    = {{IEEE}},
  year         = {2023}
}

@inproceedings{DBLP:conf/cvpr/DengGVKZ20,
  author       = {Jiankang Deng and
                  Jia Guo and
                  Evangelos Ververas and
                  Irene Kotsia and
                  Stefanos Zafeiriou},
  title        = {RetinaFace: Single-Shot Multi-Level Face Localisation in the Wild},
  booktitle    = {{CVPR}},
  pages        = {5202--5211},
  publisher    = {Computer Vision Foundation / {IEEE}},
  year         = {2020}
}

@inproceedings{an2022killing,
  title={Killing two birds with one stone: Efficient and robust training of face recognition cnns by partial fc},
  author={An, Xiang and Deng, Jiankang and Guo, Jia and Feng, Ziyong and Zhu, XuHan and Yang, Jing and Liu, Tongliang},
  booktitle={Proceedings of the IEEE/CVF conference on computer vision and pattern recognition},
  pages={4042--4051},
  year={2022}
}

@inproceedings{he2016deep,
  title={Deep residual learning for image recognition},
  author={He, Kaiming and Zhang, Xiangyu and Ren, Shaoqing and Sun, Jian},
  booktitle={Proceedings of the IEEE conference on computer vision and pattern recognition},
  pages={770--778},
  year={2016}
}

@inproceedings{deng2019lightweight,
  title={Lightweight face recognition challenge},
  author={Deng, Jiankang and Guo, Jia and Zhang, Debing and Deng, Yafeng and Lu, Xiangju and Shi, Song},
  booktitle={Proceedings of the IEEE/CVF international conference on computer vision workshops},
  pages={0--0},
  year={2019}
}

@inproceedings{fang2022unsupervised,
  title={Unsupervised face morphing attack detection via self-paced anomaly detection},
  author={Fang, Meiling and Boutros, Fadi and Damer, Naser},
  booktitle={2022 IEEE International Joint Conference on Biometrics (IJCB)},
  pages={1--11},
  year={2022},
  organization={IEEE}
}

@inproceedings{damer2022privacy,
  title={Privacy-friendly synthetic data for the development of face morphing attack detectors},
  author={Damer, Naser and L{\'o}pez, C{\'e}sar Augusto Fontanillo and Fang, Meiling and Spiller, No{\'e}mie and Pham, Minh Vu and Boutros, Fadi},
  booktitle={Proceedings of the IEEE/CVF Conference on Computer Vision and Pattern Recognition},
  pages={1606--1617},
  year={2022}
}

@inproceedings{boutros2021mixfacenets,
  title={Mixfacenets: Extremely efficient face recognition networks},
  author={Boutros, Fadi and Damer, Naser and Fang, Meiling and Kirchbuchner, Florian and Kuijper, Arjan},
  booktitle={2021 IEEE International Joint Conference on Biometrics (IJCB)},
  pages={1--8},
  year={2021},
  organization={IEEE}
}

@inproceedings{caldeira2025madprompts,
  title={MADPromptS: Unlocking Zero-Shot Morphing Attack Detection with Multiple Prompt Aggregation},
  author={Caldeira, Eduarda and Boutros, Fadi and Damer, Naser},
  booktitle={Proceedings of the 1st International Workshop \& Challenge on Subtle Visual Computing},
  pages={12--20},
  year={2025}
}

@inproceedings{radford2021learning,
  title={Learning transferable visual models from natural language supervision},
  author={Radford, Alec and Kim, Jong Wook and Hallacy, Chris and Ramesh, Aditya and Goh, Gabriel and Agarwal, Sandhini and Sastry, Girish and Askell, Amanda and Mishkin, Pamela and Clark, Jack and others},
  booktitle={International conference on machine learning},
  pages={8748--8763},
  year={2021},
  organization={PmLR}
}

@inproceedings{boutros2023cr,
  title={CR-FIQA: face image quality assessment by learning sample relative classifiability},
  author={Boutros, Fadi and Fang, Meiling and Klemt, Marcel and Fu, Biying and Damer, Naser},
  booktitle={Proceedings of the IEEE/CVF conference on computer vision and pattern recognition},
  pages={5836--5845},
  year={2023}
}

@inproceedings{DBLP:conf/icb/MazeADKMO0NACG18,
  author       = {Brianna Maze and
                  Jocelyn C. Adams and
                  James A. Duncan and
                  Nathan D. Kalka and
                  Tim Miller and
                  Charles Otto and
                  Anil K. Jain and
                  W. Tyler Niggel and
                  Janet Anderson and
                  Jordan Cheney and
                  Patrick Grother},
  title        = {{IARPA} Janus Benchmark - {C:} Face Dataset and Protocol},
  booktitle    = {{ICB}},
  pages        = {158--165},
  publisher    = {{IEEE}},
  year         = {2018}
}

@inproceedings{huber2022syn,
  title={SYN-MAD 2022: Competition on face morphing attack detection based on privacy-aware synthetic training data},
  author={Huber, Marco and Boutros, Fadi and Luu, Anh Thi and Raja, Kiran and Ramachandra, Raghavendra and Damer, Naser and Neto, Pedro C and Gon{\c{c}}alves, Tiago and Sequeira, Ana F and Cardoso, Jaime S and others},
  booktitle={2022 IEEE International Joint Conference on Biometrics (IJCB)},
  pages={1--10},
  year={2022},
  organization={IEEE}
}

@article{damer2023mordiff,
  title={Mordiff: Recognition vulnerability and attack detectability of face morphing attacks created by diffusion autoencoders},
  author={Damer, Naser and Fang, Meiling and Siebke, Patrick and Kolf, Jan Niklas and Huber, Marco and Boutros, Fadi},
  journal={arXiv preprint arXiv:2302.01843},
  year={2023}
}

@misc{ISO301073,
key = {ISO/IEC DIS 30107-3:2016},
year = {2017},
title = {{ISO/IEC DIS 30107-3:2016: Information Technology – Biometric presentation attack detection – P. 3: Testing and reporting}},
author = {{International Organization for Standardization}}
}

@article{phillips1998feret,
  title={The FERET database and evaluation procedure for face-recognition algorithms},
  author={Phillips, P Jonathon and Wechsler, Harry and Huang, Jeffery and Rauss, Patrick J},
  journal={Image and vision computing},
  volume={16},
  number={5},
  pages={295--306},
  year={1998},
  publisher={Elsevier}
}

@inproceedings{phillips2005overview,
  title={Overview of the face recognition grand challenge},
  author={Phillips, P Jonathon and Flynn, Patrick J and Scruggs, Todd and Bowyer, Kevin W and Chang, Jin and Hoffman, Kevin and Marques, Joe and Min, Jaesik and Worek, William},
  booktitle={2005 IEEE computer society conference on computer vision and pattern recognition (CVPR'05)},
  volume={1},
  pages={947--954},
  year={2005},
  organization={IEEE}
}

@article{kvanchiani2023easyportrait,
  title={EasyPortrait--Face Parsing and Portrait Segmentation Dataset},
  author={Kvanchiani, Karina and Petrova, Elizaveta and Efremyan, Karen and Sautin, Alexander and Kapitanov, Alexander},
  journal={arXiv preprint arXiv:2304.13509},
  year={2023}
}

@Article{zeng2021image,
title = {Image processing effects on the deep face recognition system},
journal = {Mathematical Biosciences and Engineering},
volume = {18},
number = {2},
pages = {1187-1200},
year = {2021},
issn = {1551-0018},
doi = {10.3934/mbe.2021064},
url = {https://www.aimspress.com/article/doi/10.3934/mbe.2021064},
author = {Jinhua Zeng and Xiulian Qiu and Shaopei Shi},
}

@article{rathgeb2019impact,
  author       = {Christian Rathgeb and
                  Antitza Dantcheva and
                  Christoph Busch},
  title        = {Impact and Detection of Facial Beautification in Face Recognition:
                  An Overview},
  journal      = {{IEEE} Access},
  volume       = {7},
  pages        = {152667--152678},
  year         = {2019},
  url          = {https://doi.org/10.1109/ACCESS.2019.2948526},
  doi          = {10.1109/ACCESS.2019.2948526},
  bibsource    = {dblp computer science bibliography, https://dblp.org}
}

@inproceedings{lee2020byeglassesgan,
  author       = {Yu{-}Hui Lee and
                  Shang{-}Hong Lai},
  editor       = {Andrea Vedaldi and
                  Horst Bischof and
                  Thomas Brox and
                  Jan{-}Michael Frahm},
  title        = {ByeGlassesGAN: Identity Preserving Eyeglasses Removal for Face Images},
  booktitle    = {Computer Vision - {ECCV} 2020 - 16th European Conference, Glasgow,
                  UK, August 23-28, 2020, Proceedings, Part {XXIX}},
  series       = {Lecture Notes in Computer Science},
  volume       = {12374},
  pages        = {243--258},
  publisher    = {Springer},
  year         = {2020},
  url          = {https://doi.org/10.1007/978-3-030-58526-6\_15},
  doi          = {10.1007/978-3-030-58526-6\_15},
  bibsource    = {dblp computer science bibliography, https://dblp.org}
}

@inproceedings{schlett2023compression,
  author       = {Torsten Schlett and
                  Sebastian Schachner and
                  Christian Rathgeb and
                  Juan E. Tapia and
                  Christoph Busch},
  title        = {Effect of Lossy Compression Algorithms on Face Image Quality and Recognition},
  booktitle    = {{IEEE} International Conference on Acoustics, Speech and Signal Processing
                  {ICASSP} 2023, Rhodes Island, Greece, June 4-10, 2023},
  pages        = {1--5},
  publisher    = {{IEEE}},
  year         = {2023},
  url          = {https://doi.org/10.1109/ICASSP49357.2023.10095832},
  doi          = {10.1109/ICASSP49357.2023.10095832},
  bibsource    = {dblp computer science bibliography, https://dblp.org}
}

@article{rathgeb2021compression,
  author       = {Christian Rathgeb and
                  Kevin Bernardo and
                  Nathania E. Haryanto and
                  Christoph Busch},
  title        = {Effects of image compression on face image manipulation detection:
                  {A} case study on facial retouching},
  journal      = {{IET} Biom.},
  volume       = {10},
  number       = {3},
  pages        = {342--355},
  year         = {2021},
  url          = {https://doi.org/10.1049/bme2.12027},
  doi          = {10.1049/BME2.12027},
  bibsource    = {dblp computer science bibliography, https://dblp.org}
}

@inproceedings{neddo2024printscanning,
  author       = {Richard E. Neddo and
                  Zander W. Blasingame and
                  Chen Liu},
  title        = {The Impact of Print-Scanning in Heterogeneous Morph Evaluation Scenarios},
  booktitle    = {{IEEE} International Joint Conference on Biometrics, {IJCB} 2024,
                  Buffalo, NY, USA, September 15-18, 2024},
  pages        = {1--10},
  publisher    = {{IEEE}},
  year         = {2024},
  url          = {https://doi.org/10.1109/IJCB62174.2024.10744441},
  doi          = {10.1109/IJCB62174.2024.10744441},
  bibsource    = {dblp computer science bibliography, https://dblp.org}
}

@inproceedings{DBLP:conf/icpr/DamerWHCB0K18,
  author       = {Naser Damer and
                  Yaza Wainakh and
                  Olaf Henniger and
                  Christian Croll and
                  Benoit Berthe and
                  Andreas Braun and
                  Arjan Kuijper},
  title        = {Deep Learning-based Face Recognition and the Robustness to Perspective
                  Distortion},
  booktitle    = {24th International Conference on Pattern Recognition, {ICPR} 2018,
                  Beijing, China, August 20-24, 2018},
  pages        = {3445--3450},
  publisher    = {{IEEE} Computer Society},
  year         = {2018},
  url          = {https://doi.org/10.1109/ICPR.2018.8545037},
  doi          = {10.1109/ICPR.2018.8545037},
  bibsource    = {dblp computer science bibliography, https://dblp.org}
}

@misc{iso19794-5_2011,
  title = {{ISO/IEC 19794-5:2011 Information technology — Biometric data interchange formats — Part 5: Face image data}},
  author = {{ISO/IEC}},
  year = {2011},
  note = {Edition 2}
}

@misc{iso20059_2025,
  title = {{ISO/IEC 20059:2025 Information technology — Methodologies to evaluate the resistance of biometric systems to morphing attacks}},
  author = {{ISO/IEC}},
  year = {2025},
  note = {Edition 1}
}

@inproceedings{DBLP:conf/icb/FerraraFM14,
  author       = {Matteo Ferrara and
                  Annalisa Franco and
                  Davide Maltoni},
  title        = {The magic passport},
  booktitle    = {{IEEE} International Joint Conference on Biometrics, Clearwater, {IJCB}
                  2014, FL, USA, September 29 - October 2, 2014},
  pages        = {1--7},
  publisher    = {{IEEE}},
  year         = {2014},
  url          = {https://doi.org/10.1109/BTAS.2014.6996240},
  doi          = {10.1109/BTAS.2014.6996240},
  bibsource    = {dblp computer science bibliography, https://dblp.org}
}

@inproceedings{DBLP:conf/visapp/MakrushinND17,
  author       = {Andrey Makrushin and
                  Tom Neubert and
                  Jana Dittmann},
  title        = {Automatic Generation and Detection of Visually Faultless Facial Morphs},
  booktitle    = {{VISIGRAPP} {(6:} {VISAPP)}},
  pages        = {39--50},
  publisher    = {SciTePress},
  year         = {2017}
}

@inproceedings{DBLP:conf/iwbf/ScherhagRB18,
  author       = {Ulrich Scherhag and
                  Christian Rathgeb and
                  Christoph Busch},
  title        = {Performance variation of morphed face image detection algorithms across
                  different datasets},
  booktitle    = {{IWBF}},
  pages        = {1--6},
  publisher    = {{IEEE}},
  year         = {2018}
}

@article{MADVgg,
  author       = {Matteo Ferrara and
                  Annalisa Franco and
                  Davide Maltoni},
  title        = {Face morphing detection in the presence of printing/scanning and heterogeneous
                  image sources},
  journal      = {{IET} Biom.},
  volume       = {10},
  number       = {3},
  pages        = {290--303},
  year         = {2021}
}

@inproceedings{DBLP:conf/cvpr/RaghavendraRVB17a,
  author       = {Ramachandra Raghavendra and
                  Kiran B. Raja and
                  Sushma Venkatesh and
                  Christoph Busch},
  title        = {Transferable Deep-CNN Features for Detecting Digital and Print-Scanned
                  Morphed Face Images},
  booktitle    = {{CVPR} Workshops},
  pages        = {1822--1830},
  publisher    = {{IEEE} Computer Society},
  year         = {2017}
}

@inproceedings{DBLP:conf/btas/RaghavendraRB16,
  author       = {Ramachandra Raghavendra and
                  Kiran B. Raja and
                  Christoph Busch},
  title        = {Detecting morphed face images},
  booktitle    = {{BTAS}},
  pages        = {1--7},
  publisher    = {{IEEE}},
  year         = {2016}
}

@book{orav2016smart,
  title={Smart borders: EU entry/exit system},
  author={Orav, Anita and D'Alfonso, Alessandro},
  year={2016},
  publisher={EPRS, European Parliamentary Research Service, Members' Research Service}
}

@misc{backgrounderaser_app,
  author       = {{Photo Editor \& Cutout Background Eraser (Developer)}},
  title        = {Photo Editor Cutout Background Eraser},
  year         = {n.d.},
  howpublished = {\url{https://play.google.com/store/apps/details?id=photoeditor.cutout.backgrounderaser}},
  note         = {Google Play Store app, accessed 2026-03-17}
}

@misc{adobe_express_bg_removal,
  author       = {{Adobe Inc.}},
  title        = {Remove Background from Image},
  year         = {n.d.},
  howpublished = {\url{https://www.adobe.com/express/feature/image/remove-background}},
  note         = {Adobe Express feature page, accessed 2026-03-17}
}

@misc{removebg,
  author       = {{Canva Austria GmbH}},
  title        = {remove.bg: Remove Image Background},
  year         = {n.d.},
  howpublished = {\url{https://www.remove.bg/}},
  note         = {Online background removal service, accessed 2026-03-17}
}

@misc{quek2019facemorpher,
  title={Facemorpher},
  author={Quek, Alyssa},
  year={2019},
  publisher={January}
}

@article{openCVmorph,
    title={Face Morph Using OpenCV — C++ / Python},
    volume={1},
    url={https://learnopencv.com/face-morph-using-opencv-cpp-python/},
    number={1},
    journal={LearnOpenCV},
    author={Satya Mallick},
    year={2016},
}

@article{debruine2018debruine,
  title={debruine/webmorph: Beta release 2},
  author={DeBruine, Lisa},
  journal={Zenodo https://doi. org/10},
  volume={5281},
  year={2018}
}

@article{DBLP:journals/corr/YiLLL14a,
  author       = {Dong Yi and
                  Zhen Lei and
                  Shengcai Liao and
                  Stan Z. Li},
  title        = {Learning Face Representation from Scratch},
  journal      = {CoRR},
  volume       = {abs/1411.7923},
  year         = {2014}
}

@inproceedings{NISTQuaity,
  author    = {P. Grother and A. Hom, M. Ngan and K. Hanaoka},
  title     = {Ongoing Face Recognition Vendor Test (FRVT) Part 5: Face Image Quality Assessment (4th Draft)},
  booktitle = {National Institute of Standards
and Technology},
  publisher = {Tech. Rep. },
  year      = {Sep. 2021},
}

@article{DBLP:journals/tbbis/ZhangVRRDB21,
  author    = {Haoyu Zhang and
               Sushma Venkatesh and
               Raghavendra Ramachandra and
               Kiran Bylappa Raja and
               Naser Damer and
               Christoph Busch},
  title     = {{MIPGAN} - Generating Strong and High Quality Morphing Attacks Using
               Identity Prior Driven {GAN}},
  journal   = {{IEEE} Trans. Biom. Behav. Identity Sci.},
  volume    = {3},
  number    = {3},
  pages     = {365--383},
  year      = {2021},
  url       = {https://doi.org/10.1109/TBIOM.2021.3072349},
  doi       = {10.1109/TBIOM.2021.3072349},
  bibsource = {dblp computer science bibliography, https://dblp.org}
}

@inproceedings{DBLP:conf/nips/SunSPT24,
  author       = {Zhonglin Sun and
                  Siyang Song and
                  Ioannis Patras and
                  Georgios Tzimiropoulos},
  title        = {CemiFace: Center-based Semi-hard Synthetic Face Generation for Face
                  Recognition},
  booktitle    = {NeurIPS},
  year         = {2024}
}

@inproceedings{DBLP:conf/iclr/KarrasALL18,
  author       = {Tero Karras and
                  Timo Aila and
                  Samuli Laine and
                  Jaakko Lehtinen},
  title        = {Progressive Growing of GANs for Improved Quality, Stability, and Variation},
  booktitle    = {{ICLR}},
  publisher    = {OpenReview.net},
  year         = {2018}
}

@inproceedings{DBLP:conf/iccv/LiuLWT15,
  author       = {Ziwei Liu and
                  Ping Luo and
                  Xiaogang Wang and
                  Xiaoou Tang},
  title        = {Deep Learning Face Attributes in the Wild},
  booktitle    = {{ICCV}},
  pages        = {3730--3738},
  publisher    = {{IEEE} Computer Society},
  year         = {2015}
}

@article{meden2021privacy,
  title={Privacy--enhancing face biometrics: A comprehensive survey},
  author={Meden, Bla{\v{z}} and Rot, Peter and Terh{\"o}rst, Philipp and Damer, Naser and Kuijper, Arjan and Scheirer, Walter J and Ross, Arun and Peer, Peter and {\v{S}}truc, Vitomir},
  journal={IEEE Transactions on Information Forensics and Security},
  volume={16},
  pages={4147--4183},
  year={2021},
  publisher={IEEE}
}

\clearpage
\section*{\textbf{SUPPLEMENTARY MATERIAL}}
\section*{SAM Prompts Selection}

\begin{figure}
    \centering
    \includegraphics[width=0.99\linewidth]{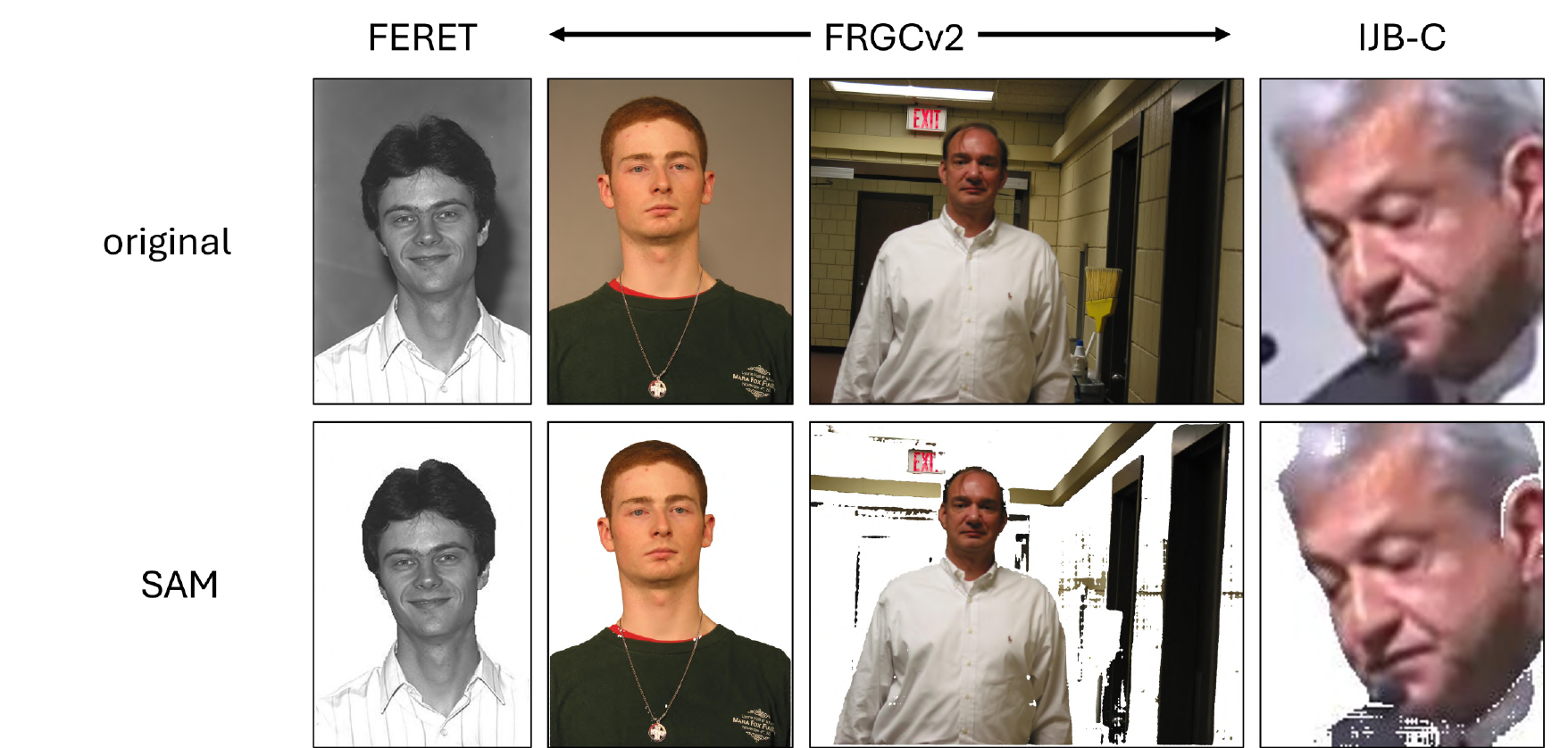}
    \caption{Pictures from FERET, FRGCv2, and IJB-C before and after background removal by SAM using the five facial landmarks and the perimeter of the box as segmentation prompts. The margins were added to highlight the images' borders after background removal. Notice that while FERET, FRGCv2 portrait, and IJB-C achieve good segmentation results, segmenting a FRGCv2 landscape picture (third column) with these prompts leads to poor background removal.}
    \label{fig:sam_ex}
\end{figure}

\textcolor{black}{As described in the main document, SAM \cite{kirillov2023segment} can receive input prompts that orient the model towards identifying the desired object. In this work, we consider two types of input prompts: a set of points describing coordinates that belong to the desired object and a bounding box that includes the desired object. The five facial landmarks extracted by RetinaFace \cite{DBLP:conf/cvpr/DengGVKZ20} define the set of points fed as the first prompt. These points are known to belong to the face and orient the segmentation process towards the inclusion of important facial regions that may be occluded (such as the eyes, when glasses or sunglasses are present). The choice of an appropriate bounding box, however, is not as straightforward due to the distinct disposition of the faces inside the image frame. Figure \ref{fig:sam_ex} shows four examples of samples segmented with SAM using the perimeter of the image as the bounding box prompt. It can be easily seen that this technique is efficient for FERET \cite{phillips1998feret}, IJB-C \cite{DBLP:conf/icb/MazeADKMO0NACG18}, and the portrait images of FRGCv2 \cite{phillips2005overview}; in these cases, the face occupies most of the frame, and using the image perimeter as a bounding box ensures that the biggest possible amount of relevant pixels is included inside the box, without significant background noise. For FRGCv2 \cite{phillips2005overview} landscape pictures, however, the background occupies a significant portion of the sample, and using the whole image perimeter as a bounding box is not discriminative enough, resulting in poor segmentation (Figure \ref{fig:sam_ex}). Hence, we define a more refined prompt for these samples by extending the bounding boxes extracted by RetinaFace \cite{DBLP:conf/cvpr/DengGVKZ20} differently in each dimension.} 

\begin{figure}
    \centering
    \includegraphics[width=0.99\linewidth]{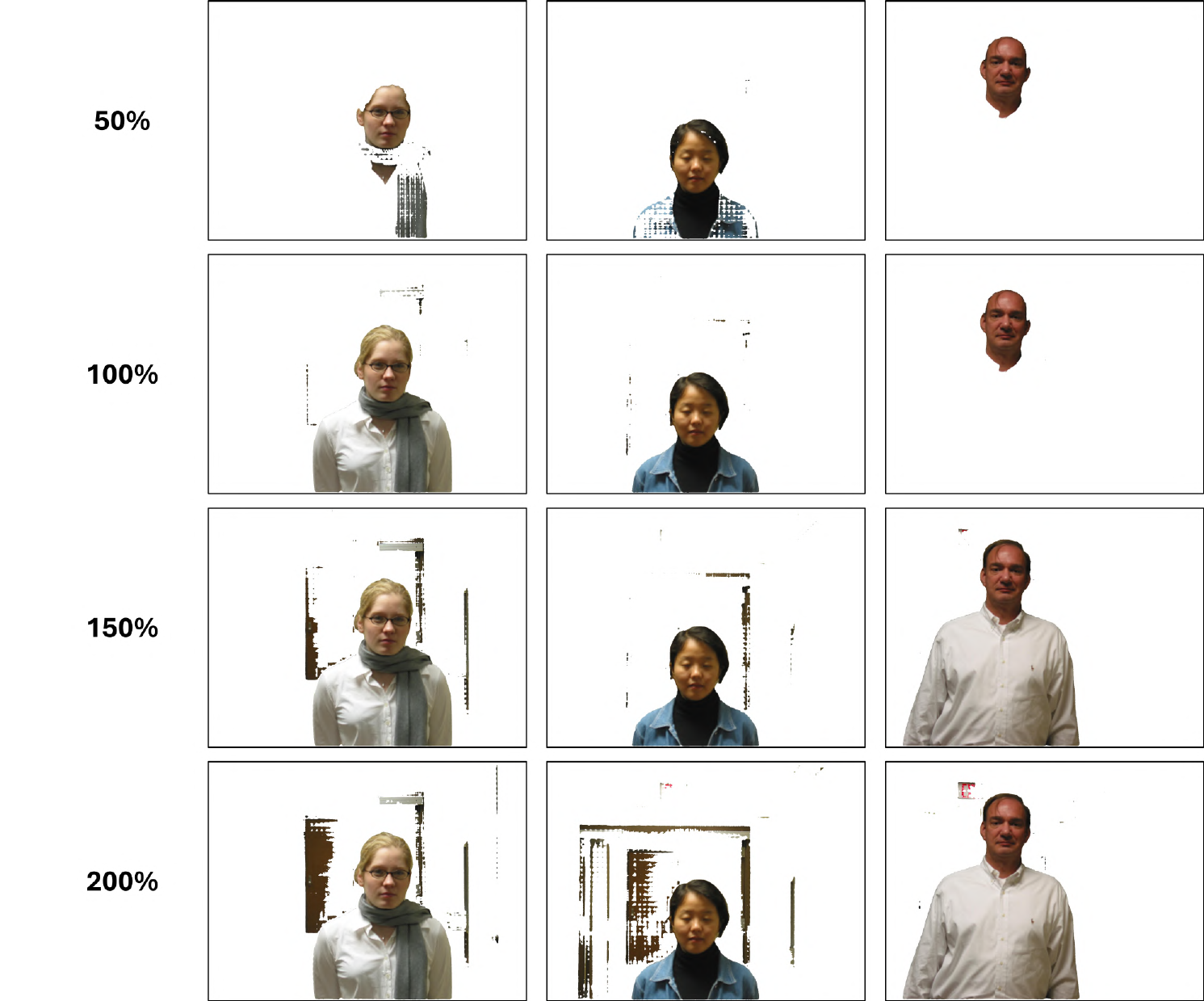}
    \caption{Landscape pictures of FRGCv2 before and after background removal by SAM using the five facial landmarks and different bidirectional horizontal extensions of the bounding box extracted by RetinaFace as segmentation prompts (the vertical limits of the box are the image up and bottom pixels). The margins were added to highlight the images' borders after background removal. Notice that a bidirectional extension of 150\% allows to segment the face, body, and accessories of the three examples.}
    \label{fig:sam_horizontal}
\end{figure}

\begin{figure}
    \centering
    \includegraphics[width=0.99\linewidth]{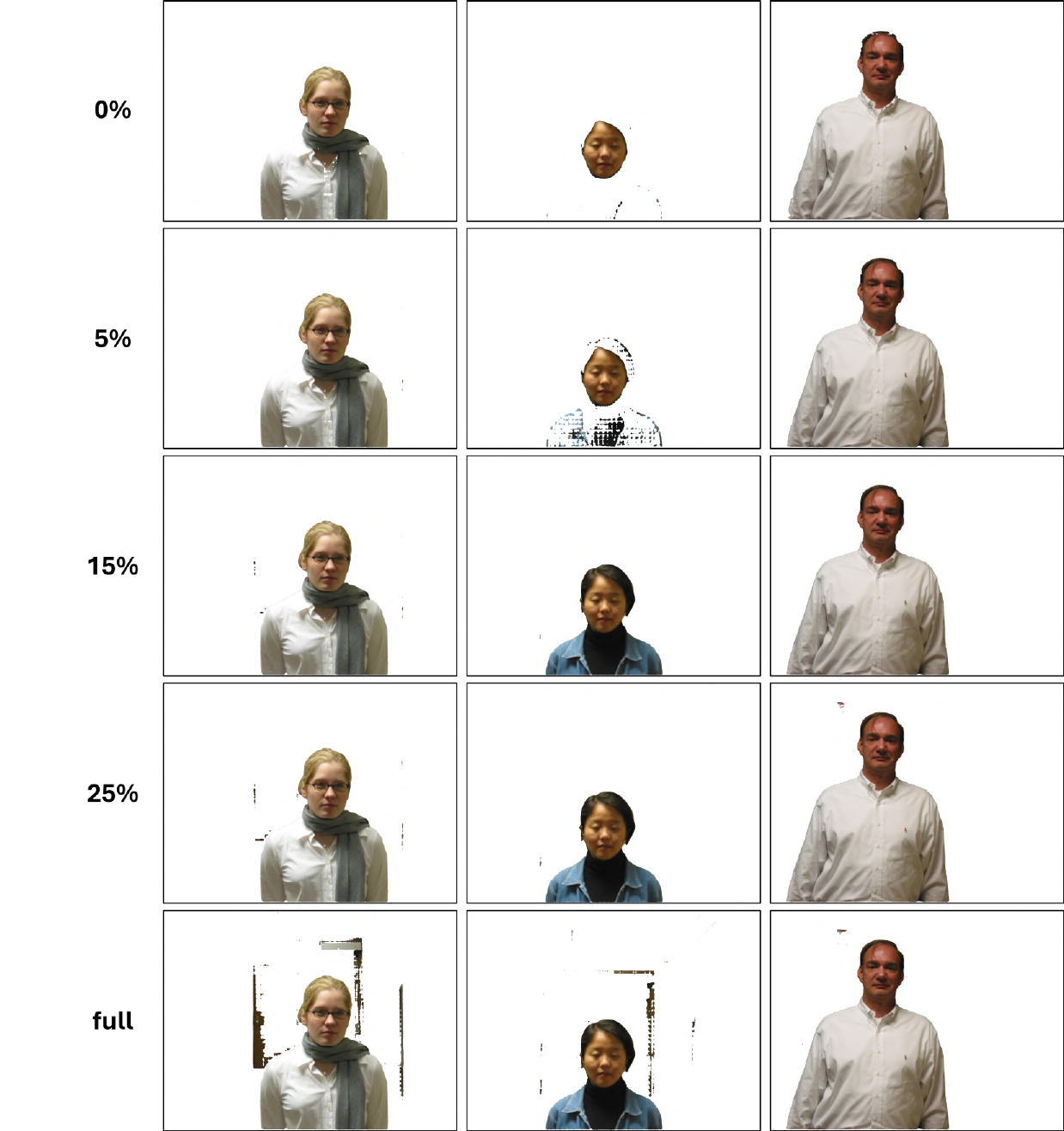}
    \caption{Landscape pictures of FRGCv2 before and after background removal by SAM using the five facial landmarks, and a bounding box with 150\% bidirectional horizontal extension and different upwards vertical extensions (the bottom limit of the box is the last row of the image). The margins were added to highlight the images' borders after background removal. Notice that a vertical upwards extension of 15\% allows to segment the face, body, and accessories of the three examples while resulting in minimal background artifacts.}
    \label{fig:sam_vertical}
\end{figure}

\textcolor{black}{We start by experimenting with the horizontal dimension by fixing the height of the prompt box as the height of the sample and widening the horizontal dimension of the RetinaFace bounding box to different extents. It is possible to observe that the best segmentation results are achieved when the bounding box is extended by 150\% in each horizontal direction, resulting in a total width of 400\% of the original size (Figure \ref{fig:sam_horizontal}). Nonetheless, it can still be seen that a significant portion of the background is not properly eliminated in these samples, due to the vertical dimensions of the prompt box. While it is logical to extend the bounding box until the bottom of the sample to capture the person's full body length, the box only needs to be minimally extended vertically. As visually represented in Figure \ref{fig:sam_vertical}, an extension of 15\% allows to keep the face, body and hair, while efficiently eliminating most of the background.}

\section*{Genuine and Impostor Score Distributions}

\begin{figure*}
    \centering
    \includegraphics[width=0.77\linewidth]{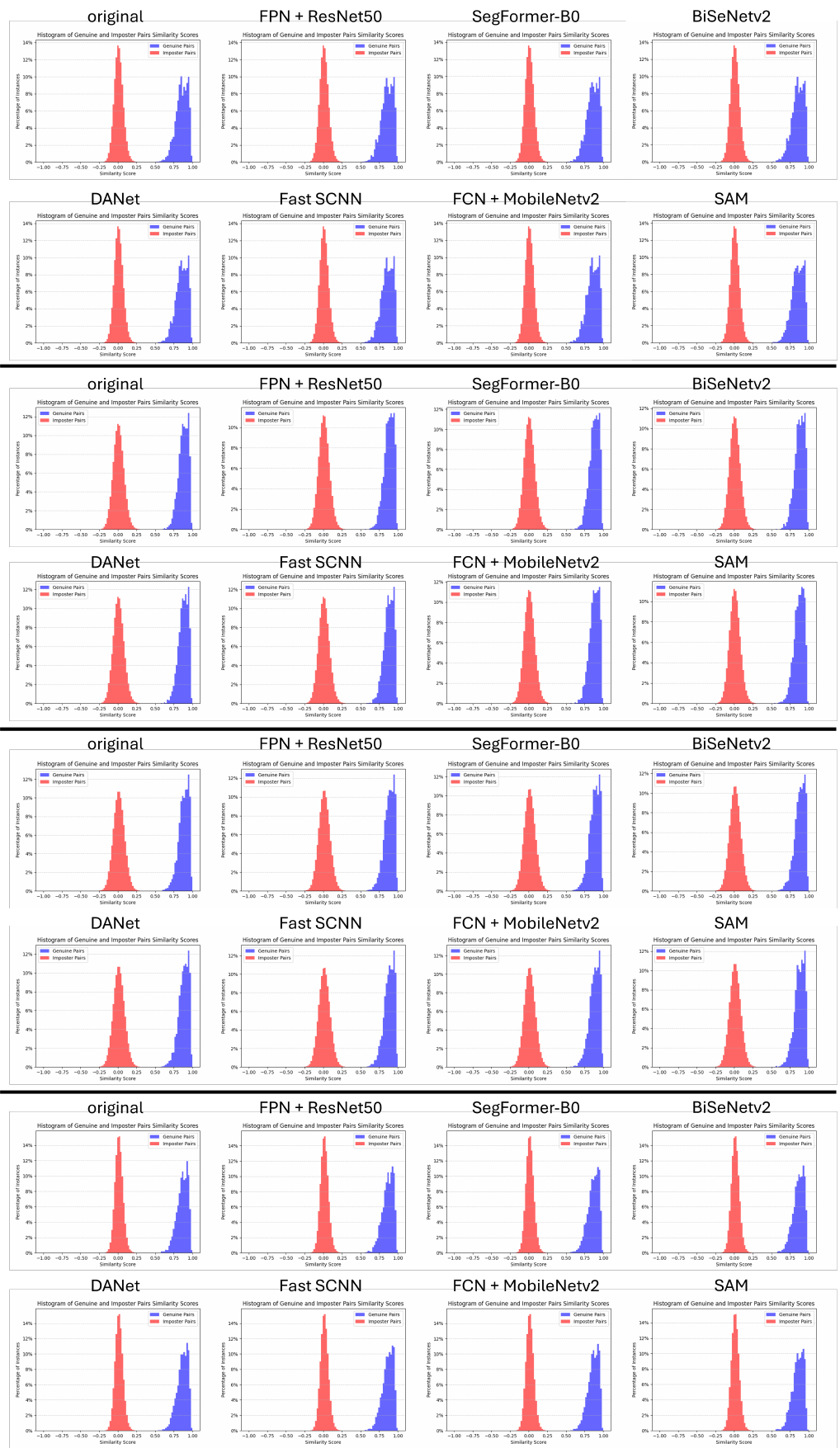}
    \vspace{-3mm}
    \caption{Histograms of the genuine and impostor score distributions of FERET and its segmented versions when evaluated by ElasticFace (first block), ArcFace (second block), SwinFace (third block), and TransFace (fourth block). The absence of visually perceptible differences between the graphs indicates that segmenting images acquired under controlled conditions does not significantly impact FR performance.}
    \label{fig:feret_dist}
\end{figure*}

\begin{figure*}
    \centering
    \includegraphics[width=0.77\linewidth]{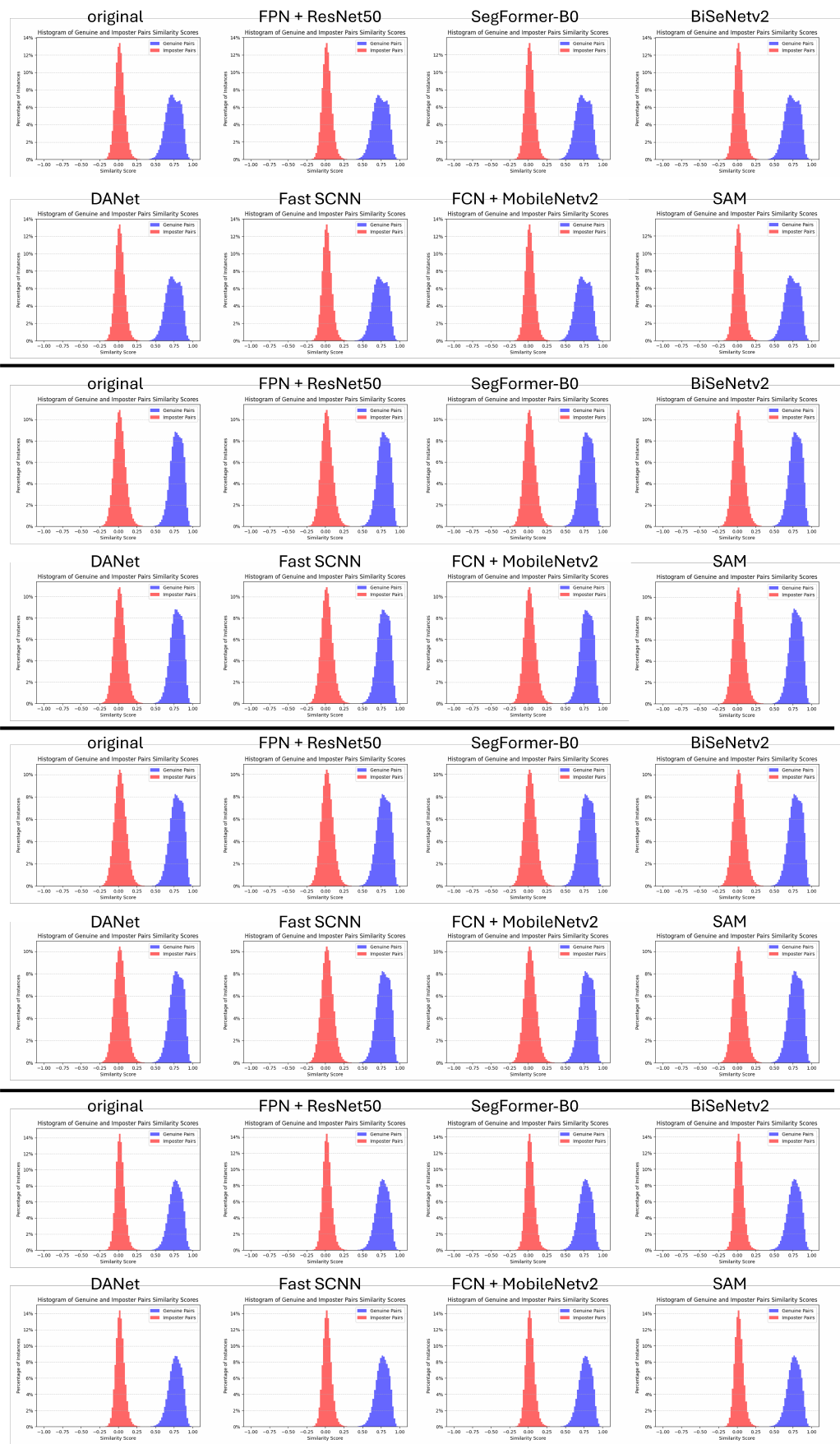}
    \vspace{-3mm}
    \caption{Histograms of the genuine and impostor score distributions of FRGCv2 and its segmented versions when evaluated by ElasticFace (first block), ArcFace (second block), SwinFace (third block), and TransFace (fourth block). The absence of visually perceptible differences between the graphs indicates that segmenting images acquired under the semi-controlled conditions of FRGCv2 does not significantly impact FR performance.}
    \label{fig:frgc_dist}
\end{figure*}

\begin{figure*}
    \centering
    \includegraphics[width=0.77\linewidth]{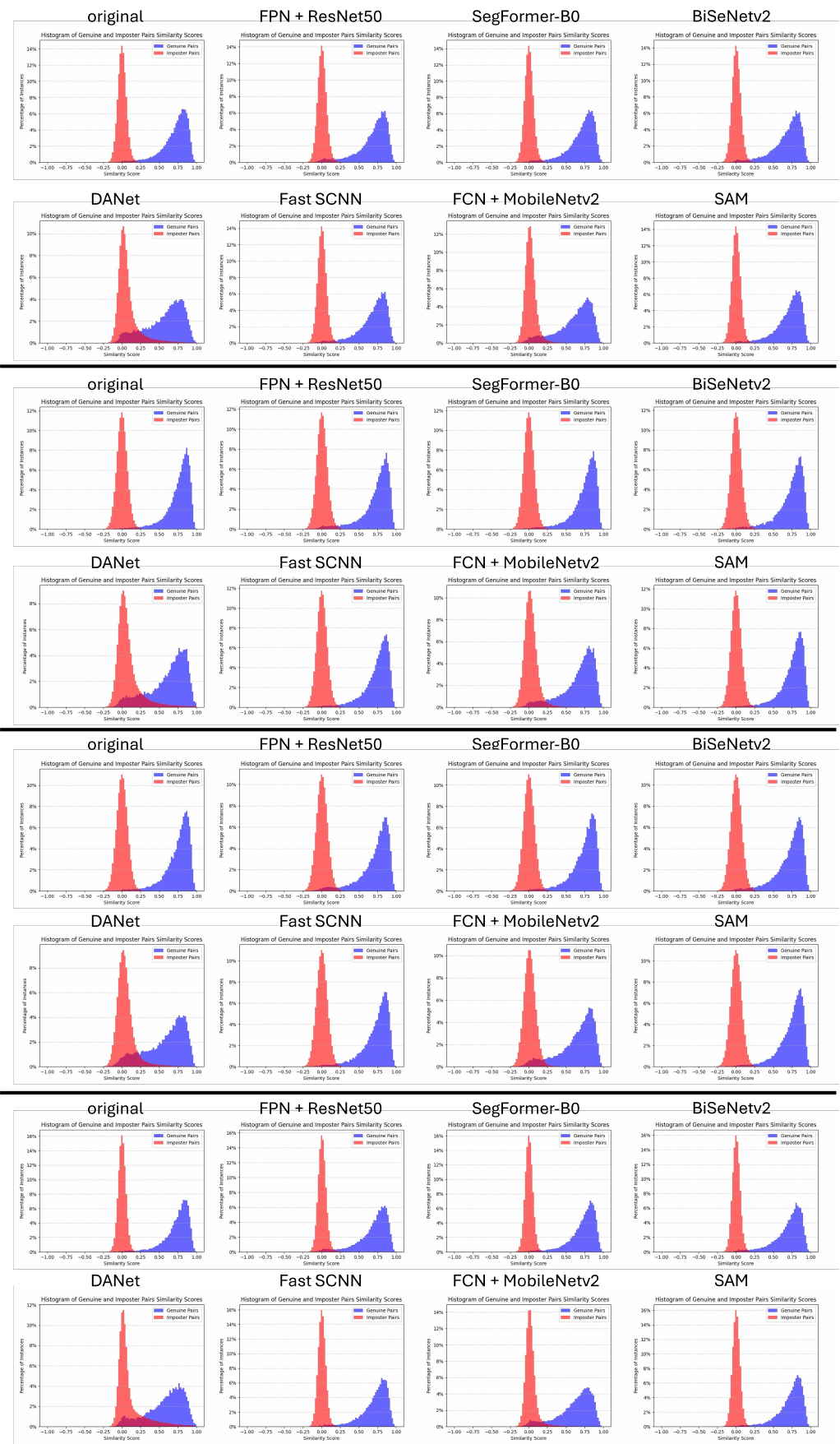}
    \vspace{-3mm}
    \caption{Histograms of the genuine and impostor score distributions of IJB-C and its segmented versions when evaluated by ElasticFace (first block), ArcFace (second block), SwinFace (third block), and TransFace (fourth block). The significant differences in genuine/impostor distributions' overlap between segmentation methods directly translate to the performance discrepancies verified in FR evaluation.}
    \label{fig:ijbc_dist}
\end{figure*}

\textcolor{black}{Figures \ref{fig:feret_dist}, \ref{fig:frgc_dist} and \ref{fig:ijbc_dist} show the histograms of the genuine and impostor score distributions of FERET \cite{phillips1998feret}, FRGCv2 \cite{phillips2005overview}, IJB-C \cite{DBLP:conf/icb/MazeADKMO0NACG18} and their segmented variants, when evaluated by the four FR models considered in the main paper (ElasticFace \cite{DBLP:conf/cvpr/BoutrosDKK22}, ArcFace \cite{DBLP:conf/cvpr/DengGXZ19}, SwinFace \cite{DBLP:journals/tcsv/QinWDWCHD24} and TransFace \cite{DBLP:conf/iccv/DanLXD0XS23}). It is possible to observe that there are no significant differences between the plots obtained by the same FR model when analyzing each version of FERET (Figure \ref{fig:feret_dist}). Despite being hard to notice visually in the histograms, it should be kept in mind that these differences exist due to a slight approximation between the who score distributions for segmented datasets, as discussed in the main document. However, due to the small magnitude of these approximations, the clear separation between the genuine and impostor distributions is maintained regardless of the selected segmentation techniques, which justifies the absence of FR performance drops after segmentation, as reported in the main document. Similar conclusions can also be drawn regarding FRGCv2 (Figure \ref{fig:frgc_dist}). For IJB-C and its segmented variants (Figure \ref{fig:ijbc_dist}), the differences between the distributions are significant. While the histograms of SegFormer-B0 \cite{xie2021segformer} and SAM \cite{kirillov2023segment} are similar to the original distribution, segmentation techniques such as DANet \cite{fu2019dual} and FCN+MobileNet \cite{long2015fully} lead to a significantly larger overlap between genuine and impostor distributions. These results directly align with those of the main paper, since segmenting with these networks resulted in extremely poor FR performance, accompanied by a significant drop in average face image quality and in the difference between the average of genuine and impostor scores, $\Delta$.}

\section*{Joint Metrics' Visualization}
\begin{figure*}
    \centering
    \includegraphics[width=0.99\linewidth]{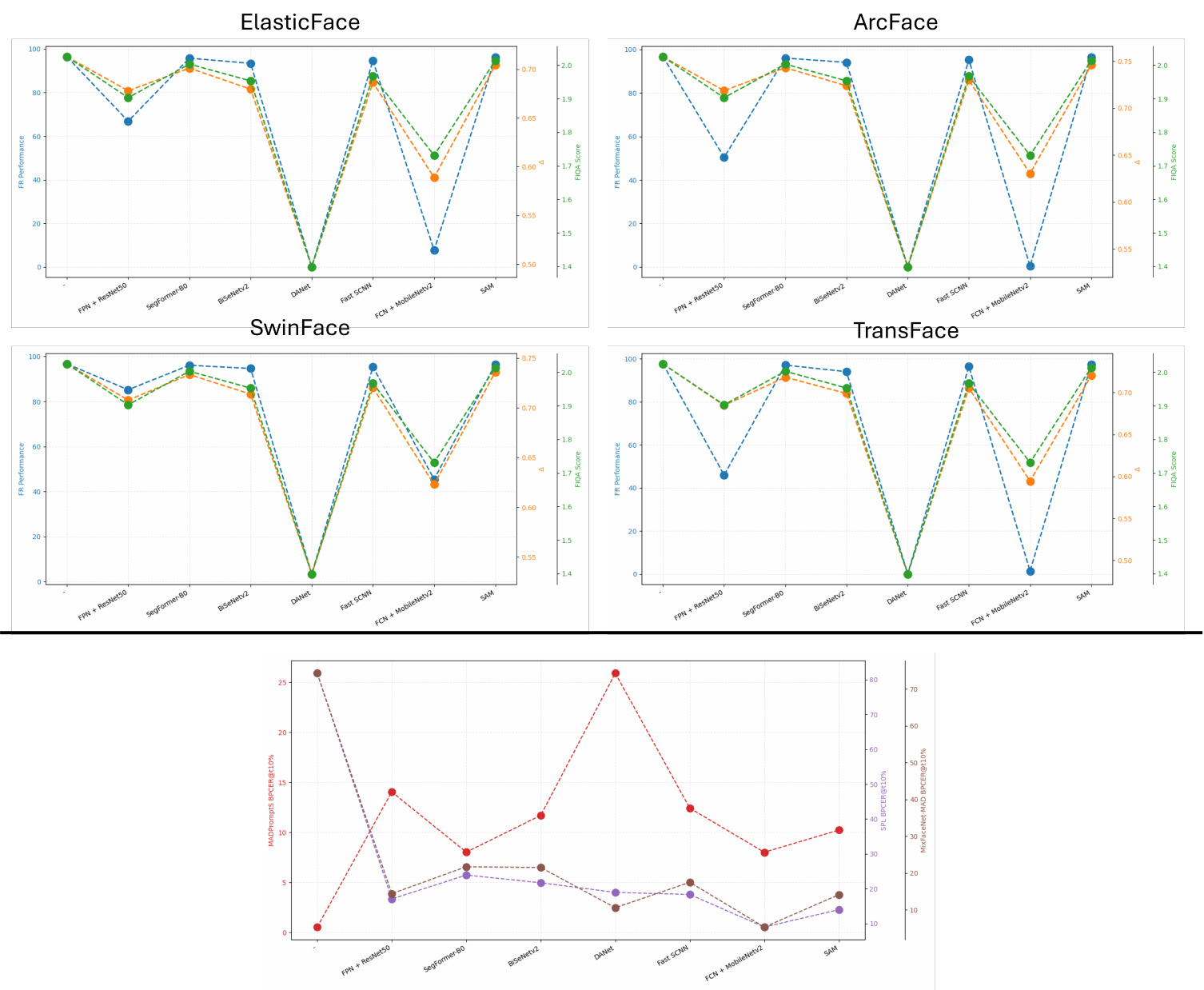}
    \vspace{-3mm}
    \caption{Joint visualization of the metrics evaluated for IJB-C and its segmented variants in the main paper, namely TAR@FAR=1e-4, $\Delta$, the FIQA metric (top block), and BPCER@t10\% for the three evaluated MAD methods (bottom block).}
    \label{fig:correlations}
\end{figure*}

\textcolor{black}{To further emphasize the correlation between the different extracted metrics and to more easily visualize whether their variation tendency matches, we show all of them together in the same plot. In particular, we plot a graph showing the values TAR@FAR=1e-4, $\Delta$ and the FIQA metric for IJB-C and its segmented variants. We include an extra graph with the values of BPCER@t10\% achieved by the three considered MAD methods. We do not include this analysis for FERET and FRGCv2 due to the absence of significant differences in FR performance and image quality after segmentation for these datasets. As it can be easily seen, the FR performance, $\Delta$, and image quality values follow a similar tendency regardless of the network used to evaluate the considered dataset (Figure \ref{fig:correlations}). This behavior is not surprising given the connection between these three metrics, with lower face image quality and higher overlap between genuine and impostor scores distributions (lower $\Delta$) being usually associated with poorer FR performance. Regarding the MAD evaluation, the results of MADPromptS \cite{caldeira2025madprompts} also follow a similar tendency, with datasets that achieve higher quality / FR performance triggering the MAD systems wrongly less often (lower BPCER@t10\%). For SPL \cite{fang2022unsupervised} and MixFaceNet-MAD \cite{damer2022privacy}, however, datasets with a poorer recognition performance are generally less prone to trigger erroneous detection. As analyzed in the main document, these results show that different MAD systems present significantly different behaviors when evaluating the same segmented data. Hence, background removal can prove beneficial or impairing depending on the considered MAD system.}

\section*{Segmentation Methods Evaluation}

\textcolor{black}{When evaluating how background removal techniques affect downstream tasks such as FR and MAD, it is important to keep in mind that the verified performance changes might derive from inappropriate background removal. While it is impossible to ensure perfect segmentation, the quality of the employed segmentation techniques should be enforced in advance to ensure that verified performance fluctuations can be mapped to the act of background removal, and not to the specific technique being employed. In this section, we describe the experimental setup and results of this evaluation.}

\subsection{Evaluation Dataset}
\textcolor{black}{We evaluate the seven background removal techniques considered in this work on an independent database, namely the CelebAMask-HQ test subset \cite{DBLP:conf/nips/SunSPT24}. CelebMask-HQ mask is a high-quality dataset labeled according to CelebA-HQ \cite{DBLP:conf/iclr/KarrasALL18}, which contains 30,000 high-resolution face images from CelebA \cite{DBLP:conf/iccv/LiuLWT15}. Each image is accompanied by a series of ground truth (GT) binary segmentation masks for 19 facial elements \cite{DBLP:conf/nips/SunSPT24} (left/right brow, left/right eye, upper/lower lip, ...). We start by determining the union of these masks to recover the GT mask for the complete face, resulting in 2,824 test images with their corresponding GT face segmentation. Since the segmentation protocol followed for SAM requires the coordinates of the facial landmarks, we exclude the only image for which RetinaFace \cite{DBLP:conf/cvpr/DengGVKZ20} was not able to detect any landmarks, resulting in 2,823 test images.}

\subsection{Evaluation Metrics}

\textcolor{black}{Let the true positives, false positives, true negatives, and false negatives be defined as TP, FP, TN, and FN, respectively. To evaluate whether the face is correctly segmented, we determine the mean intersection over union of the detected object ($mIoU$) and the recall of the prediction. The recall measures the segmentation method's ability to find all positive instances, and is defined as the ratio between the TP and the ground truth positives:}

\begin{equation}
    Recall = \frac{TP}{TP + FN}.
\end{equation}

\textcolor{black}{The $mIoU$ is defined as the ratio between the intersection and the union of the predicted and ground truth masks:}

\begin{equation}
    mIoU = \frac{TP}{TP+FP+FN}.
\end{equation}

\textcolor{black}{However, it is important to keep in mind that metrics such as the recall and $mIoU$ hide segmentation failures when a significant portion of the background is considered as belonging to the object. This is particularly true for recall, but $mIoU$ is also associated with this phenomenon when the object occupies a significant portion of the image (FP is bounded to a low value if TP is high), as happens on the CelebAMask-HQ dataset (Figure \ref{fig:seg_eval}). This poor segmentation will also be hidden when evaluating the FR performance on the segmented samples, as inappropriate background removal approximates the samples to their original unsegmented versions. Hence, we also report %the mean intersection over union of the background and 
the precision, which measures the accuracy of positive predictions, that is, how many predicted positives are actual positives:}
%The metrics are defined by Equations \ref{eq:miou_bkgrd} and \ref{eq:precision}, respectively.}

\begin{equation}
\label{eq:precision}
    Precision =\frac{TP}{TP + FP}.
\end{equation}

\subsection{Results}

\begin{figure}
    \centering
    \includegraphics[width=0.99\linewidth]{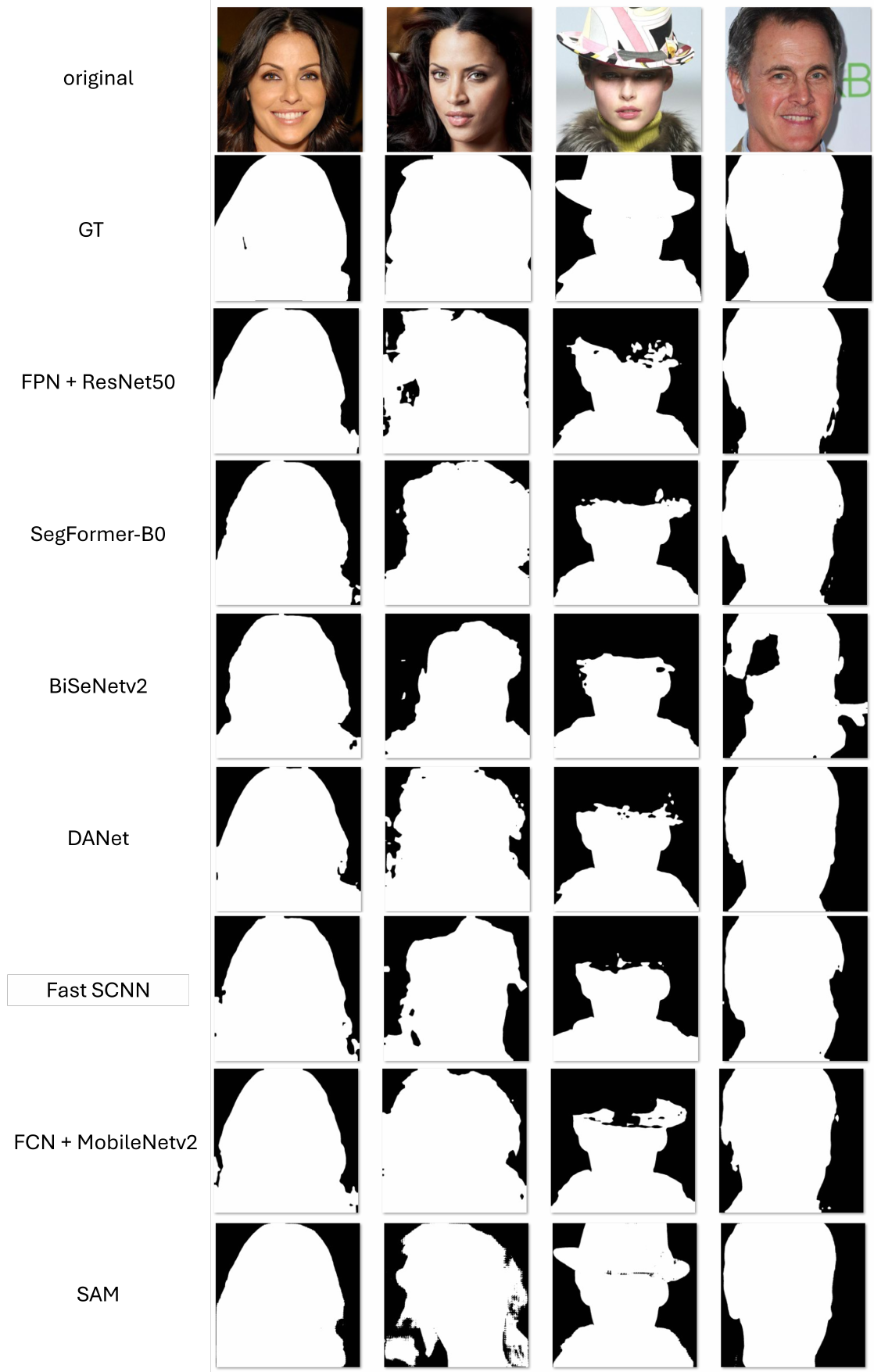}
    \vspace{-3mm}
    \caption{Segmentation masks obtained by each of the considered segmentation methods in four CelebAMask-HQ samples. The second row depicts the GT segmentation masks.}
    \label{fig:seg_eval}
\end{figure}

\textcolor{black}{Table \ref{tab:segmentation_eval} shows the segmentation evaluation results of the seven methods used for background removal in this work. While most methods achieve values of $mIoU$, precision, and recall above 90\%, BiSeNetv2 and Fast SCNN achieve significantly lower $mIoU$ and recall values than the remaining methods, suggesting that they possess a lower capacity to correctly identify the desired object. This tendency can be visually confirmed in Figure \ref{fig:seg_eval}, where it is shown that these networks struggle to correctly segment some samples. However, it should be kept in mind that these results are not directly translatable to what is verified with other datasets, namely those considered in the main paper, due to their significantly different characteristics. In particular, it should be noted that the images contained in CelebAMask-HQ are high-quality samples that do not possess significant background noise, as the face occupies most of the sample. While these results do not directly translate to the tendencies observed in the main paper, there is no other face segmentation evaluation dataset with more challenging samples whose masks include clothes and accessories, to the extent of our knowledge. When looking at harder samples, such as the third image in Figure \ref{fig:seg_eval}, it is much clearer that there is a significant difference between the evaluated networks' capacity to deal with occlusions and appropriately segment accessories, with SAM being the only model capable of including the hat in the output mask.}

\begin{table}[t]
    \centering
    \caption{Segmentation evaluation of the seven methods used for background removal. The $mIoU$ and the recall evaluate how well the object is segmented in the resulting mask, while precision highlights whether the method successfully identifies background pixels as not belonging to the object.}
    \resizebox{0.99\linewidth}{!}{%
    \begin{tabular}{c|ccc}
    \hline
     \multirow{2}{*}{\textbf{Segmentation Method}} & \multicolumn{3}{c}{\textbf{Segmentation Evaluation (\%)}} \\ 
    & $mIoU$ & Recall  & Precision \\ \hline 
    FPN + ResNet50 \cite{lin2017feature} & 93.14 & 95.89 & 96.97 \\ 
    SegFormer-B0 \cite{xie2021segformer} & 93.01 & 96.07 & 96.64 \\ 
    BiSeNetv2 \cite{yu2018bisenet} & 84.41	& 88.43 & 94.91 \\ 
    DANet \cite{fu2019dual} & 92.61 & 95.91  & 96.37 \\
    Fast SCNN \cite{DBLP:conf/bmvc/PoudelLC19} & 84.64 & 88.89	 & 94.72\\ 
    FCN + MobileNetv2 \cite{long2015fully} & 90.67	 & 94.83 & 93.67\\
    SAM \cite{kirillov2023segment} & 92.41	& 94.03 & 97.88 \\ \hline 
    \end{tabular}}
    \label{tab:segmentation_eval}
\end{table}

\end{document}